\documentclass[preprint,12pt]{elsarticle}

\usepackage[usenames,dvipsnames]{color}

\usepackage[colorlinks=true,linkcolor=RoyalBlue,citecolor=RoyalBlue,urlcolor=RoyalBlue]{hyperref}

\usepackage{booktabs}
\newcommand{\tabincell}[2]{\begin{tabular}{@{}#1@{}}#2\end{tabular}}
\usepackage{textcomp,marvosym}
\usepackage{amsmath,amssymb}
\usepackage{changepage}
\usepackage[utf8x]{inputenc}
\usepackage{nameref,}
\usepackage{lineno}
\usepackage{microtype}
\DisableLigatures[f]{encoding = *, family = * }
\usepackage[table]{xcolor}
\usepackage{array}
\newcolumntype{+}{!{\vrule width 2pt}}

\newlength\savedwidth

\usepackage[aboveskip=1pt,labelfont=bf,labelsep=period,justification=raggedright,singlelinecheck=off]{caption}

\makeatletter
\renewcommand{\@biblabel}[1]{\quad#1.}
\makeatother

\definecolor{mygray}{gray}{.9}
\usepackage{graphicx}
\usepackage{amssymb}

\usepackage{float}

\usepackage{changepage}

\journal{ArXiv.org. } 

\begin{document}

\begin{frontmatter}


\title{Learning scale-variant features for robust \quad\\ iris authentication with \quad\\ deep learning based ensemble framework}



\author{\sffamily Siming Zheng\textsuperscript{1\Yinyang*},
Rahmita Wirza O.K. Rahmat\textsuperscript{2\ddag},
Fatimah Khalid\textsuperscript{3\ddag},
Nurul Amelina Nasharuddin\textsuperscript{4\ddag}
}

\address{\textbf{\sffamily \textsuperscript 1} {\sffamily Computer Assisted Surgery and Diagnostic (CASD), Department of Multimedia, Putra Malaysia University, 43400 UPM Serdang, Malaysia}
\\
\textbf{\sffamily \textsuperscript 2} {\sffamily C1,Department of Multimedia, Putra Malaysia University, 43400 UPM Serdang, Malaysia}
\\
\textbf{\sffamily \textsuperscript 3} {\sffamily C2,Department of Multimedia, Putra Malaysia University, 43400 UPM Serdang, Malaysia}
\\
\textbf{\sffamily \textsuperscript 4} {\sffamily C2,Department of Multimedia, Putra Malaysia University, 43400 UPM Serdang, Malaysia}}

\begin{abstract}
In recent years, mobile Internet has accelerated the proliferation of smart mobile development. The mobile payment, mobile security and privacy protection have become the focus of widespread attention. Iris recognition evolves into a high-security authentication technology in these fields, and widely used in distinct science fields in biometric authentication researches. The  Convolutional Neural Network (CNN) is one of the conventional deep learning approach for image recognition, whereas its anti-noise ability is weak and needs a certain amount of memory to train in image classification tasks. Under these conditions we improved the architecture of Mask R-CNN and put forward the fine-tuning neural network architectures based on mobile Inception V4, which integrate every component in an overall system that combines the iris detection, extraction, and recognition function as an iris recognition system. The proposed framework has the characteristics of scalability and high availability; it not only can learn the scale-variant features by the zero-padding normalization but also enhancing the robustness of the whole learning framework. Importantly, our custom architectures can be trained by using different spectrum of samples, such as Visible Wavelength (VW) and Near Infrared (NIR) iris biometric image data. The recognition average accuracy of 99.10\% is achieved while executing in the mobile edge calculation device of the Nvidia Jetson Nano.
\end{abstract}


\end{frontmatter}

\noindent
\Yinyang These authors contributed equally to this work.
\\
\ddag These authors also contributed equally to this work.
\\
*Corresponding author: Siming Zheng\textsuperscript{1\Yinyang*}
\\
*E-mail: \href{mailto:casdmedical@upm.edu.my}{casdmedical@upm.edu.my}
\\
*The first version was at arXiv:1912.00756v1, [v1] Mon, 2 Dec 2019
\\
*This is the second revision, [v2] Sat, 13 June 2020


\section{\rmfamily Introduction}
\label{S:1}

The essential character of iris informatization is the digitalization and recessiveness. On the one hand, the iris is one of the complex organs of the human body, the hidden password in the eyeball is richer, and contains much more random texture patterns than using the Personal Identification Number (PIN), fingerprint, and the human face \cite{ref1} \cite{ref2}. Under these favorable conditions, iris characteristics can provide its value for identifying encryption technology. On the other hand, more and more iris recognition technologies have been applied to mobile devices, such as smart mobile phones, tablets, and human-machine interactive devices, due to the GPU with high-performance graphics processing capabilities \cite{ref3} \cite{ref4} \cite{ref5}. 

Over the last two decades, smart mobile devices have been embedded with built-in high-resolution imaging sensors. The sensor can be used to perform iris recognition tasks and allow researchers to explore appropriate solutions to finish all the detailed stages of iris recognition in a mobile environment. Some iris authentication functions are supported in earlier mobile devices. For example, the first smartphone with an iris authentication in the world called Arrowsnxf-04G \cite{ref6}, which is equipped with an infrared camera and a light-emitting diode (LED), the camera can scan and decode the images of user iris. To support the iris authentication development, Samsung S8 series mobile phones added Infrared Radiation (IR) and iris camera in its front lens, using the front camera assists the iris camera with infrared LED to determine the approximate general outline of the user. The iris camera scans the iris information through the light source and then converts the iris information into a specific code. Finally, the system compares the code with a known password to determine to unlock or not. Huawei company introduced GPU Turbo technology in 2019, which can greatly energize the performance of the graphics processing on several smartphones P30 Pro series \cite{ref7}, up to 60\%. In addition, Nvidia Shield tablet K1 \cite{ref8} equipped Kepler architecture with 192 cores streaming graphics multiprocessors, which supports thousands of threads to implement high-performance calculations in parallel. 

Although the processing power of these mobile devices is growing, the system robustness is still the most concerned about. Currently, there are two research problems have to be addressed in a complex mobile environment \cite{ref9} \cite{ref10}. In this research, the first problem is how to process and recognize the high quality images of iris with scale-variant features, because most of the mobile phones are equipped with a high-definition camera and always places in an uncooperative environment. The second problem relates to how to improve the recognition accuracy on target iris images under the different spectrum for applying in different practical scenarios scenarios. Therefore, our research objective is to investigate the structure of the multi-learning model to solve the research problems above in mobile environment.  

The remainder of this paper is organized as follows: We briefly introduce related works about iris authentication in Section II. Section III puts forward the proposed iris authentication framework with multiple critical components for iris region extraction and matching. Section IV shows experimental findings in favor of using the proposed method and the analysis in Section V. Finally, we discuss and conclude the research in Section VI and VII. 
\quad\\

\section{\rmfamily Related Works}
\label{S:1}
Iris authentication is a process of identifying individuals based on the iris shape and texture distribution. We have reviewed the strengths and weaknesses of current iris authentication studies. As a result of the survey, several studies have shown that some scholars used similarity computation methods for measuring the similarity between the two iris templates, as posited in \cite{ref11} \cite{ref12} \cite{ref13} \cite{ref14} \cite{ref15} and got a lot of valuable conclusions. 

The most critical work of the iris verification system is to detect the iris and outer boundary correctly. In the study \cite{ref16} of Deshpande et al., they used Daugman's integral and differential algorithm \cite{ref17} in their experiment. Once the iris boundary is detected, the program converts the iris to a standard size and encodes it in the iris template for matching between the iris templates. The Rubber Sheet model is used for the normalization of the iris image. In the final matching phase, applying a 1D logarithmic Gabor filter for the iris feature extraction, while the Hamming distance is used as a matching algorithm to compare two biometric templates for iris verification. One of the challenges in seeking the boundaries is the iris images with low contrast or low lighting during the detection, Deshpande's algorithms overcome some difficulties in the non-uniform illumination as well as the reflections. His works enhance the performance of the segmentation and normalization process in iris authentication systems and achieve an overall accuracy of 95\% with robust characteristics.

In Mohammed Hamzah Abed’s work \cite{ref18}, he adopted Circular Hough transform to detect an iris in the recognition system with a precision rate of 98.73\% on average. And then the Haar wavelet transforms extracted the raw features from iris images for fast computing and low storage. The method of Principal Component Analysis (PCA) was to alleviate a variety of noises for image reduction and improved discriminative information for matching. In the verify phase, the method of Cosine distance measured the similarity between two non-zero vectors of inner product space, the result of the experiment shows that the method is effective with a classification accuracy of 91.14\%. Subsequent works have continued to use the same experimental data as a way to explore the performance of the proposed framework. Kaudki \cite{ref19} introduced a new thought of iris preprocessing, integrated the method of Rubber-Sheet Unwrapping and Haar wavelet transforms into every experimental object. Rubber-Sheet Unwrapping Normalization has the ability to deal with different sizes of extracted images due to the changes in pupil size caused by external lighting. The processing of normalization mainly improves the clarity of acquired features. Furthermore, the Haar wavelet transform is applied for feature extraction because of its computational simplicity. The Hamming distance, as a measure of the characteristic distance, is used to conduct and validate the target data. The recognition accuracy does not continue to increase substantially, and it remains at around 97\%. By analysis, recognizing the high-quality iris images is the weakness of the traditional template matching method. Iris images processed in a series of transformations suffer from significant degradation \cite{ref20}, making iris recognition between the training set and testing set with less relevance. It might be due to the image degradation by some image fusion techniques \cite{ref21} \cite{ref22} \cite{ref23}, and feature extraction operation like Hough transform \cite{ref24} \cite{ref25} \cite{ref26}. These findings reinforce the importance of researching the influence of image quality. 

To measure the robustness, Bansal proposed a fuzzy filtering-based approach in the iris recognition system \cite{ref27}. It consists of fuzzy noise detection, noise filtration, normalization, data enhancement, and feature extraction operations. All of sub-modules have been proposed to clean the inputted iris image before they are subjected to iris recognition. The merit of the proposed approach is high recognition rate of 99.75\%. However, the call chain is too complicated in the program that it causes some problems like module design complexity (iv(G)) in a mobile environment \cite{ref28}. The high iv(G) means that the module has a high degree of coupling, making the module difficult to isolate, maintain, and reuse. Importantly, such complex computation process leads to an increment in the CPU utilization, which translates into throughput gains in the mobile terminal. How to improve the module structure has become a big problem for researchers to be faced with \cite{ref29} \cite{ref30}.   

In a comprehensive study of iris recognition, Chiara Galdi et al. \cite{ref31} found that the combination of different classifiers describing various factors of the iris can get an ideal accuracy in iris recognition. In the color descriptor, the comparison of Euclidean distance is computed between the two images based on the color histograms, and the texture descriptor calculated a number of features by using the Minkowski Bouligand dimension \cite{ref32} \cite{ref33} through a process of decomposition. Furthermore, the cluster descriptor provided important information, such as centroid coordinates, orientation, and eccentricity. When comparing two iris images, the algorithm matches the list of two cluster feature vectors according to the all-versus-all scheme and calculates the average distance of each matching pair. All descriptors are implemented via a weighted sum for recognizing the iris. Finally, the best matching with a distance value is obtained from each matching pair, and the final weight value can predict and determine whether the iris images observed belong to the same category or not. The studies show that there are still existing problems like low recognition rate in the images with different resolutions due to insufficient feature acquisition by the classifiers \cite{ref34} \cite{ref35} \cite{ref36}.  

As was mentioned in the previous reviews, researchers must focus on not only the performance for processing and recognizing high-resolution iris data but also the storage space for operation data in mobile environments. Moreover, the systems of iris authentication on mobile devices should take a short time to decide whether a person is real or an impostor.
\quad\\
\section{\rmfamily Material and methods}
\label{S:1}
A robust iris authentication system includes detection, segmentation, normalization, and recognition submodule \cite{ref37}. Every submodule is sequentially performed in the Graphics Processing Unit (GPU) in our experiments.  Fig 1 below shows the flow of pre-processing required for the training model. We added two kinds of layers, extraction and normalization, as the robust components, which are implemented respectively in our proposed model. By doing this, the normalized feature is better for the iris recognition phase. 
\begin{figure}[!h]
\includegraphics[width=13.8cm]{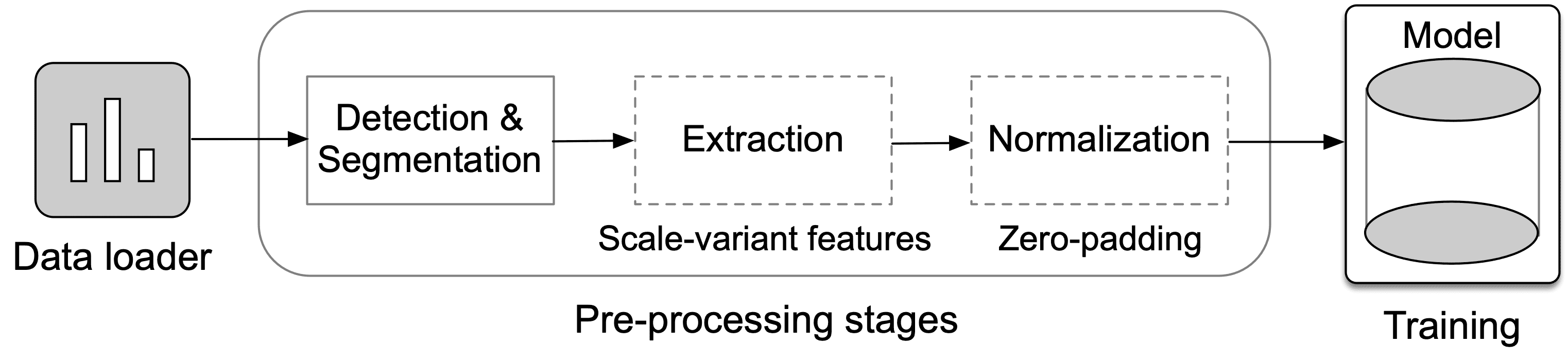}
\begin{flushleft}{\mbox{\bf Fig 1. The flow of pre-processing.} The flowchart shows every step that need to be performed before iris recognition model (From \textit{left} to \textit{right}). Notice that we propose the custom sub-components of extraction and normalization, which are both within the scope of the research for extracting and improving the scale-variant iris features in our learning model architecture. See the boxes with dotted border area. }
\rule[0.25\baselineskip]{\textwidth}{0.3pt}
\end{flushleft}
\label{fig}
\end{figure}

\subsection{The Improved Mask R-CNN Architecture}
The preprocessing of iris automatic detection is a crucial stage in the iris authentication system. We analyzed that some potential factors may be susceptible to interference, which could affect recognition performance \cite{ref38}. For instance, the natural iris texture can be easily obscured by cosmetic contacts, and this has a significant impact on the segmentation of the iris area. The experimental dataset consists of thousands of iris images taken from different angles and under various conditions are affected by a variety of internal and external factors.

\begin{figure}[!h]
\includegraphics[width=13cm]{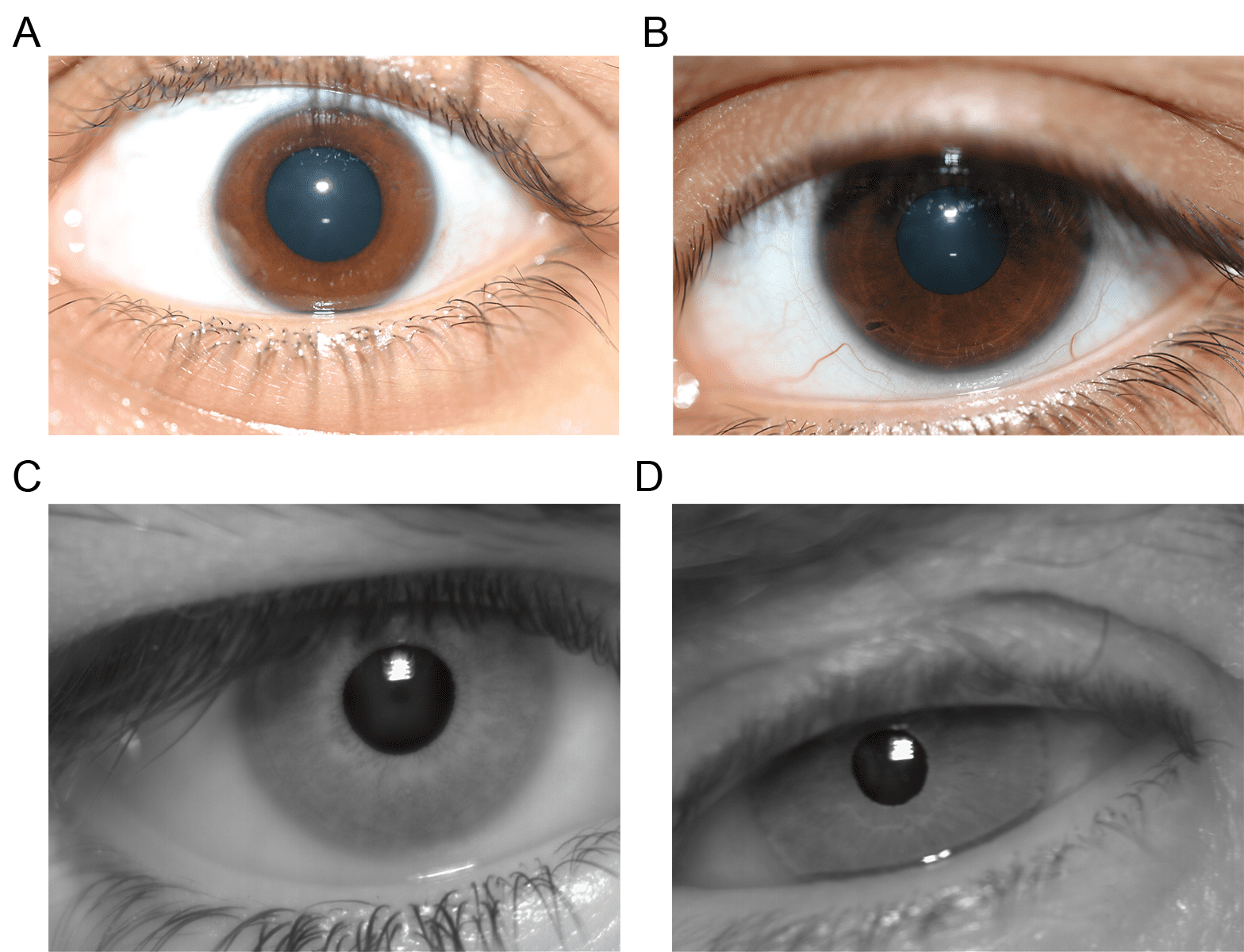}
\begin{flushleft}{\mbox{\bf Fig 2. The challenges of UTiris datasets in a different spectrum.} The color iris images (\textit{A} and \textit{B}) are acquired in the \textit{VW} and acquired grey iris images (\textit{C} and \textit{D}) under the \textit{NIR}. The images were taken from \cite{ref44}.}
\rule[0.25\baselineskip]{\textwidth}{0.3pt}
\end{flushleft}
\label{fig}
\end{figure}

As can be seen from example A in Fig 2, a slender specular highlight is shown in the iris area near the nose. Thus, the performance is reduced owing to the light reflection or uneven illumination in the area of iris. Example B showing a permeable contact lens covered in the iris region, this can generate some large artifacts in the detection region. Example C of iris image showing a small portion of eyelashes that make a considerable influence on structural iris texture. Example D of iris image exhibiting non-ideal iris region due to poor coordination of human-machine interaction, this causes many iris features to be lost. The challenges we described in the detection and extraction phases, from wearing the contact lens to the eyelashes occlusion, to specular highlights of illumination, to uncooperative action. Besides, the shape of the eyelids varies from one individual to another. All of these factors make the localization of the eyelids more difficult \cite{ref39} \cite{ref40} \cite{ref41}.     

To detect iris area at a fine-grained level, we custom the Mask R-CNN \cite{ref42} neural network architecture. The model of Mask R-CNN is an extended structure of Faster R-CNN \cite{ref43}, including a function of pixel-to-pixel alignment. The proposed Mask R-CNN is constructed by stacking the layer for predicting the iris location in each Region of Interest (ROI), called a mask branch, this layer is similar to the existing boundary layer and classification layer. Our Mask R-CNN architecture mainly performs two different operations: object detection and semantic segmentation, which are used for the instance segmentation base on iris datasets, and their aims to extract and normalize the iris region from the object images. 
\quad\\
\subsubsection{The RPN Component}
Region Proposal Networks (RPN) is an essential component of Mask R-CNN. It is a lightweight neural network that used to replace the selective search in the model of Faster-RCNN. Similar to Faster R-CNN, the purpose of RPN is to seek and generate the region proposals effectively. Besides, RPN supports calculations on the GPU to improve processing speed and back-propagation by different sizes. 

\begin{figure}[!h]
\quad\\
\centerline{\includegraphics[width=10cm]{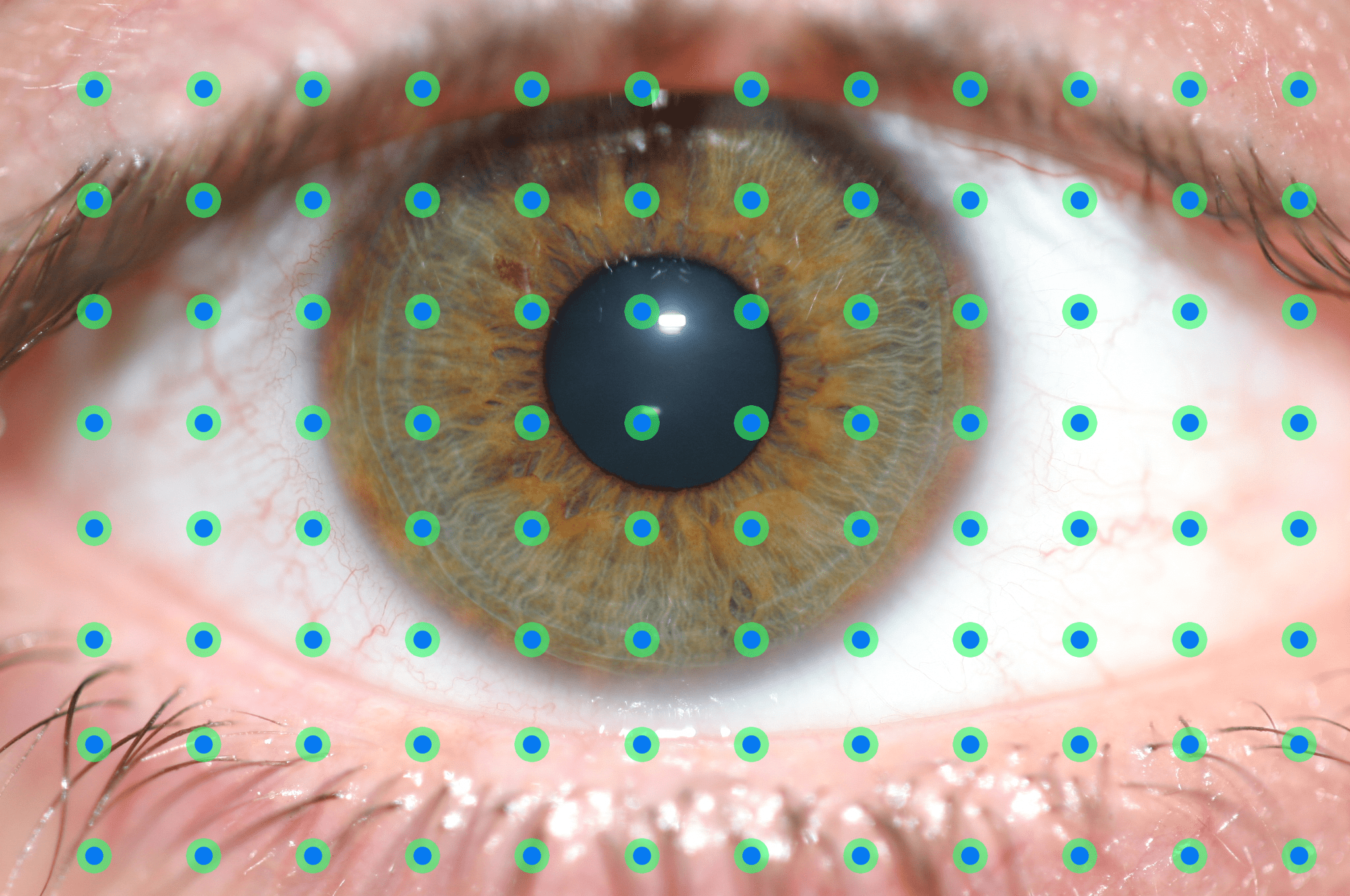}}
\begin{flushleft}{\textbf{Fig 3. The generated anchor points distributed on the raw iris image.} The figure above visualizes the effect of anchor distribution. The number of anchors is obtained through the RPN network, which is used to select positive and negative samples and finally calculate the difference between the positive sample and the ground truths \cite{ref42}.}

\rule[0.25\baselineskip]{\textwidth}{0.3pt}
\end{flushleft}
\label{fig}
\end{figure}

In the RPN architecture, a sliding window is used for scanning the image and finding the area where the target of the iris exists. It is a rectangle distributed over the image area, as demonstrated. The center of every sliding window is the anchor as shown in Fig 3, and every anchor is implicated in the aspect ratio and the scale \cite{ref42} \cite{ref43}. The sliding windows are implemented by the RPN convolution, and they scan the surrounding area based on the anchor points at high speed. Besides, sliding windows scan all areas in parallel mode by using the GPU and complete the operation within 10ms.

By default, we used three aspect ratios and three scales, resulting in 9 anchors at each sliding position. For example, for the size of W×H convolutional feature map, the 3x3 convolution networks slide on W×H by the padding operation and the stride is 1, meaning that the sliding window has W×H slides, and each slide has 9 anchors in each scanning operation. Finally, the number of W×H×9 anchors are generated, the role of RPN is to use these anchors to determine the location of the feature map and the size of its bounding box. 

At each sliding window position, multiple scanned regions are considered as candidates, which are predicted simultaneously. In our setting, a limited number of the highest potential candidate regions for each iris image is counted as k boxes. In the region layer of Mask-RCNN, the outputs of 4k parameters are needed to encode the coordinates of 4 different points for the k boxes, and the classification layer requires 2k values to evaluate the probability that each region is the target or not.   

\quad
\begin{figure}[!h]
\centerline{\includegraphics[width=12cm]{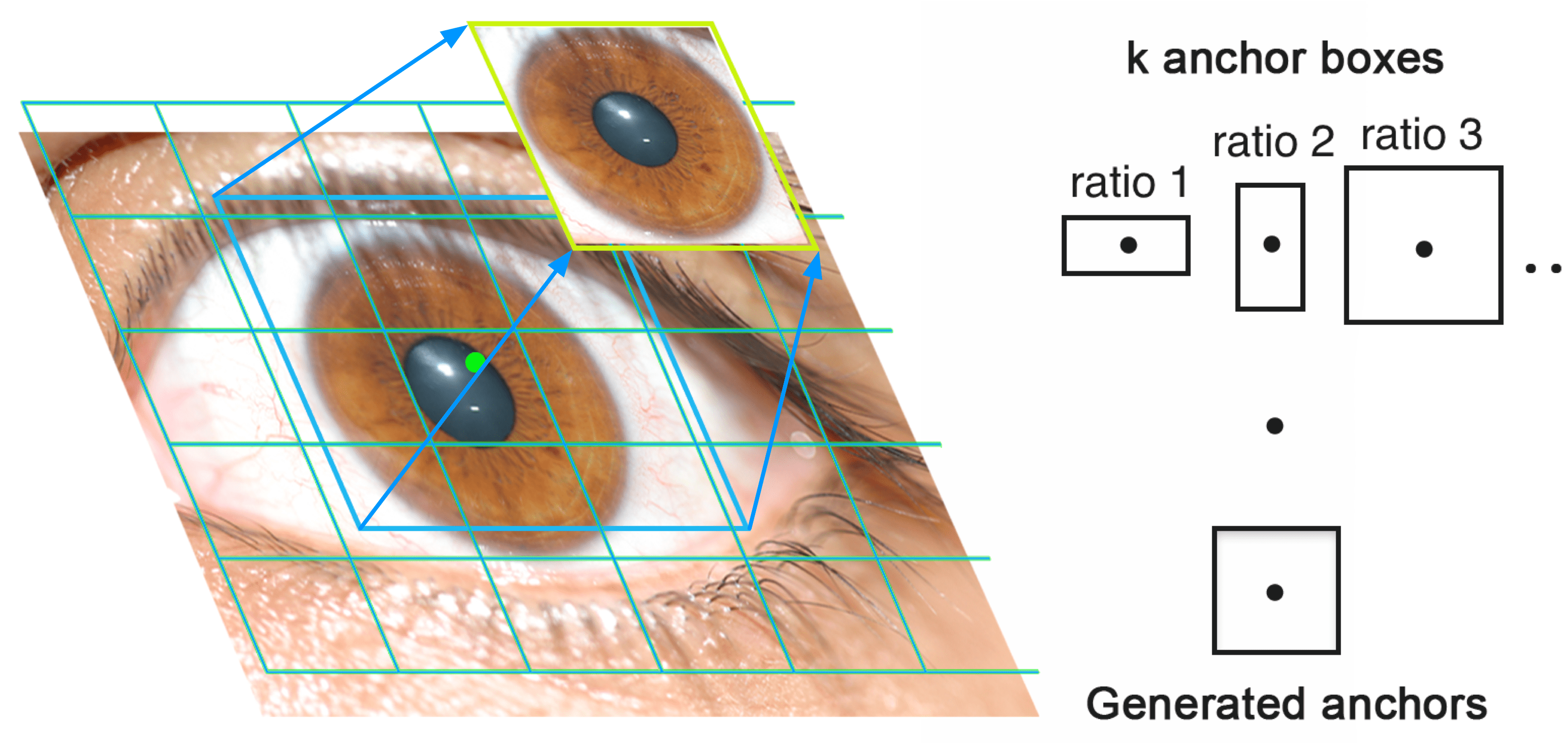}}
\mbox{\bf }  
\begin{flushleft}{\mbox{\bf Fig 4.The processing of iris region extraction.} The processing of iris region extraction. The figure demonstrates how the iris localization and iris region extraction, which are performed to provide discrimination information into a neural network in detail. The scale and aspect ratio of the anchor is controlled by the \textit{“SCALES”} and \textit{“RATIOS”} parameters in configuration.}
\rule[0.25\baselineskip]{\textwidth}{0.3pt}
\end{flushleft}

\label{fig}
\end{figure}

The Fig 4 above illustrates the processing to seek and exact the ROI of iris. Using the sliding window and anchors, we obtained W×H×9 proposals from one original iris image. Next, each proposal generated six parameters: two parameters (\textit{0} and \textit{1}) are used to label the foreground and the background probability, which is to calculate the target by comparing each proposal and ground truth; Meanwhile, each proposal is transformed into ground truth size by the translation and contraction operation. In this processing, it needs four parameters (\textit{upper-left, upper-right, lower left and lower right}) for the locating four coordinates since the differences in the position and size of each proposal and ground truth. Once six parameters are appropriately set, all the scale-variant features of the detected region can be output, including the various coordinates of iris detection. Besides, this enables us to implement further normalization function by using the extracted iris image with the highest probability.

\quad
\subsubsection{The Instance Segmentation on Experimental Data}
The experimental iris images not only contain the regions of interest (ROI) of iris but also existing redundant identifying information, such as eyelashes and sclera, etc. Thus, the current raw iris images cannot be used directly in the training model. We focused on iris preprocessing and recognition implements in the robustness and practicability of the system. The crucial ability of our framework is to extract unique properties from the iris images, and this makes it easier to create a specific code of an iris for everyone. In our experiment, the reconstructed Mask R-CNN is used for locating and extracting the iris with scale-variant features. Before inputting unique scale-variant iris features and texture into the recognition model, two steps need to be performed in our robust iris authentication system, including the iris extraction and normalization.

In the iris region detection, we propose efficient detection method specifically for the iris region, limbic and boundary of the eyelid on our dataset. To collect some training-testing samples, we create some ground truth in the dataset of UTiris for building iris ground truth. The experimental data is a set of color and grey iris images with high-resolution 2048×1360 or 1000×776 pixels. To begin with, we randomly select 158 iris images from UTiris dataset \cite{ref44}, and each user has two images for training the Mask R-CNN model. The generated image is labeled as 1 if there is evidence of the iris in the sliding window of 299×299 pixel or 0 background, as presented in Fig 5.    

\begin{figure}[!h]
\centerline{\includegraphics[width=12cm]{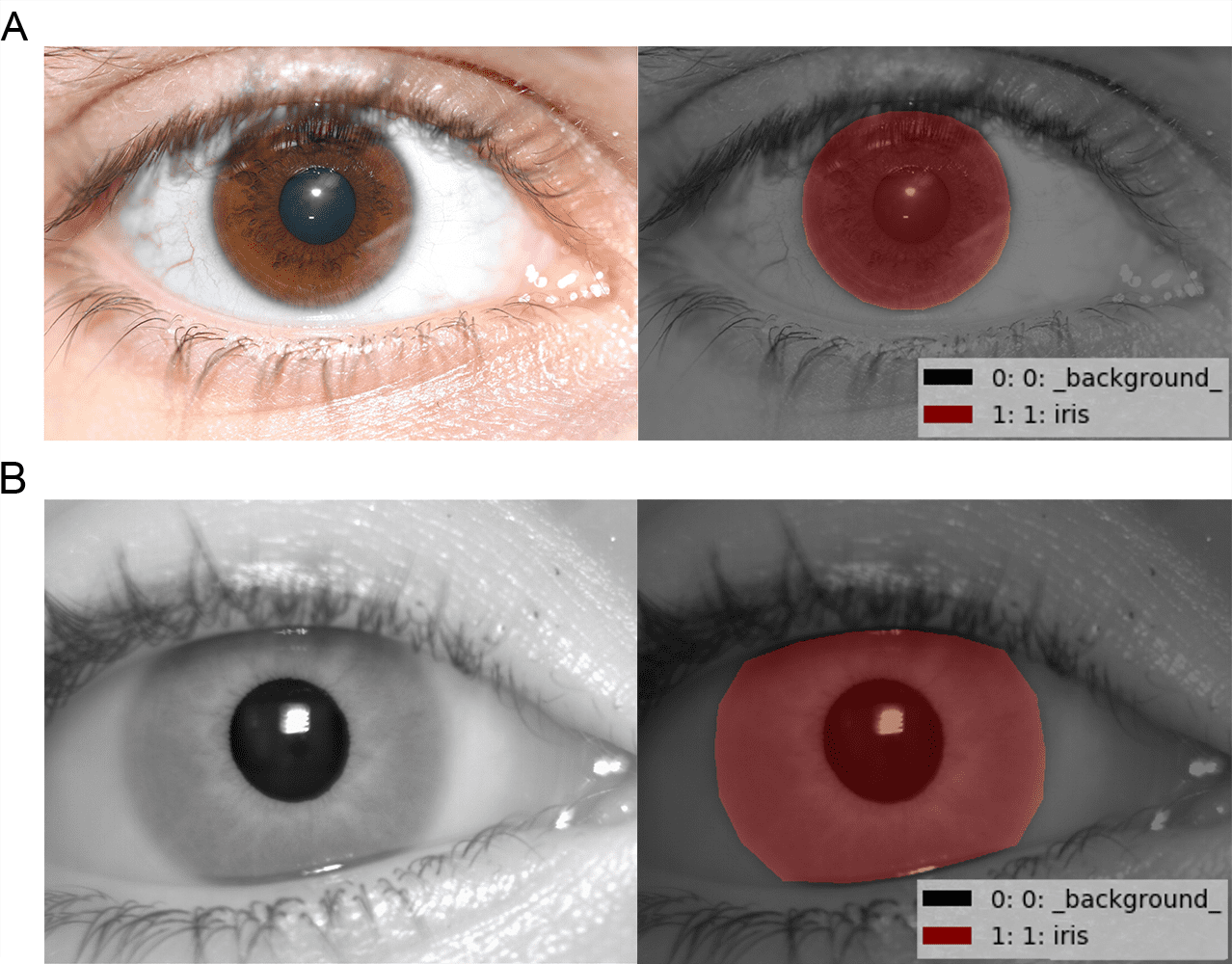}}
\mbox{\bf }  
\begin{flushleft}{\textbf{\bf Fig 5. The annotation on the fine-grained iris images.}Corresponding pairs of iris ground truth acquired in the \textit{VW} and \textit{NIR} session.  The figure given at the \textit{left side} is original iris dataset and the \textit{right side} is iris ground truth layer with red region. }
\rule[0.25\baselineskip]{\textwidth}{0.3pt}
\end{flushleft}
\label{fig}
\end{figure}

Then, to enhance the robustness of the model input, we generated a multi-channel space that duplicates the same array from grey space to a representative grey image. The new arrays formed by stacking the given arrays in three dimensions. We adopt the pre-trained model train on the COCO dataset \cite{ref45}. By matching the testing set and ground truths, the proposed Mask-RCNN calculates their matched probability to output the best one, which are intended for referencing position to locate the iris region of the raw iris images, as indicated in Fig 6(B). Ground truths were matched across different iris examples can be seen in (\href{https://figshare.com/articles/The_iris_matching_based_on_ground_truths/10280492}{DOI 10.6084/m9.figshare.10280492}). If the raw iris image was successfully matched by the top probability of iris ground truth, which can generate four coordinates. They correspond to four corners of a square in Fig 6(C). The iris position can be routed through the squares with different sizes, the results of iris localization displayed in Fig 6(D), the experimental results show good localization performance (\href{https://figshare.com/articles/Localization_performance_of_UTiris/10280501}{DOI 10.6084/m9.figshare.10280501}). After the iris is located, the iris outer boundary may contain some periocular textures, such as sclera, eyelid and pupil, which can contribute to further analysis and correlation \cite{ref46}. In addition, all of these textures also play a significant role that is expected to complement iris as auxiliary features for improving the recognition effect in non-cooperative environments \cite{ref47} \cite{ref48}. The same calculation mode is applied for the grey iris images arraying from Fig 6(E) to 6(H).
\begin{figure}[!ht]
\centerline{\includegraphics[width=13.8cm]{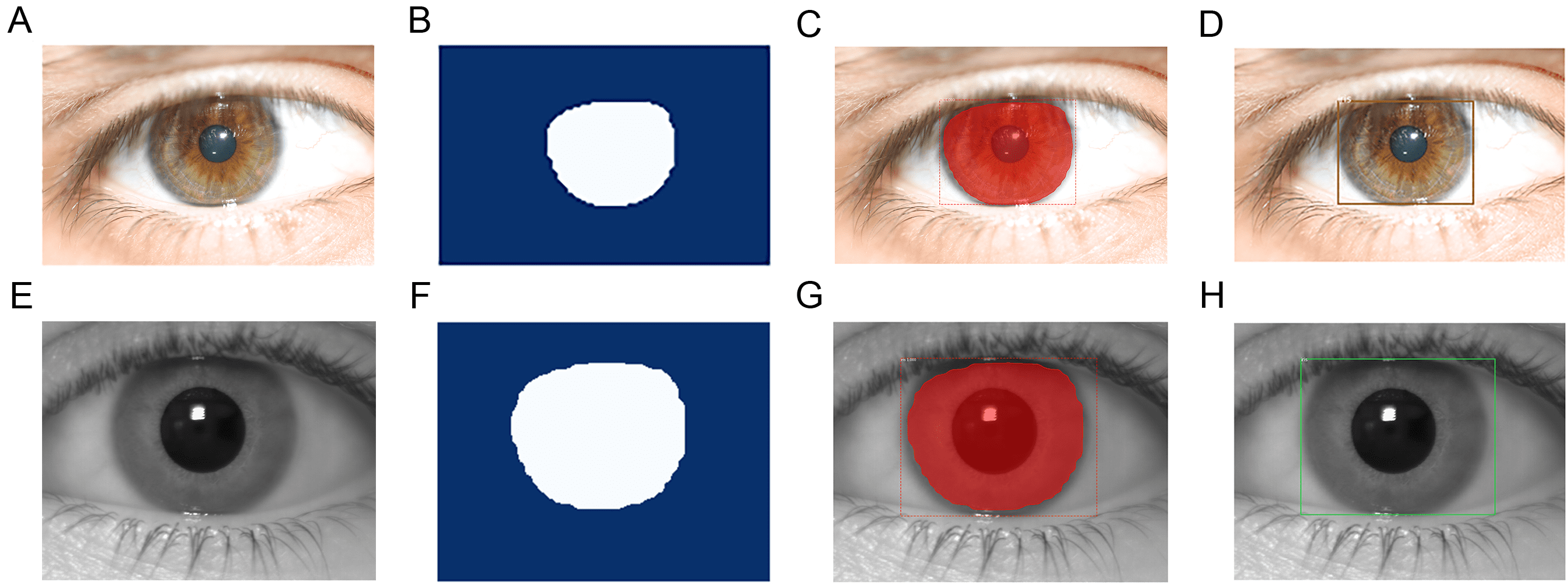}}
\mbox{\bf }  
\begin{flushleft}{\mbox{\bf Fig 6. The iris localization with iris ground truths.} To achieve iris region localization, three normalized steps corresponding to the top iris ground truth (\textit{2nd column}), iris region (\textit{3rd column}), iris periocular (\textit{4th column}), are obtained under the different spectrum images applied by the function of proposed Mask-RCNN instance segmentation. The \textit{upper} row in each sub-diagram shows successful iris region segmentation in the \textit{VW} session, and the \textit{bottom} row shows the segmentation in the \textit{NIR} 
session. }
\rule[0.25\baselineskip]{\textwidth}{0.3pt}
\end{flushleft}
\label{fig}
\end{figure}

\subsection{\normalsize The Zero-padding Normalization on the Scale-variant Features}
During the testing recognition model, the fully connected layer needs to reshape the pooled results after the convolutional layers, so the input images of the fully connected layer need to be a set of fixed size images. If the dimension of the input vector is not fixed, then the number of the weight parameter of the full connection is variable, which results in slow testing and dynamic changes of the network, and has almost no effect on parameter training. In the previous step, all of extracted iris region images are the same wide size of 299 pixels, but all images are diverse in height size (less than 299 pixels). To fill these vacant pixels is an effective way for enhancing and training the recognition model.  

\begin{figure}[!ht]
\includegraphics[width=14cm]{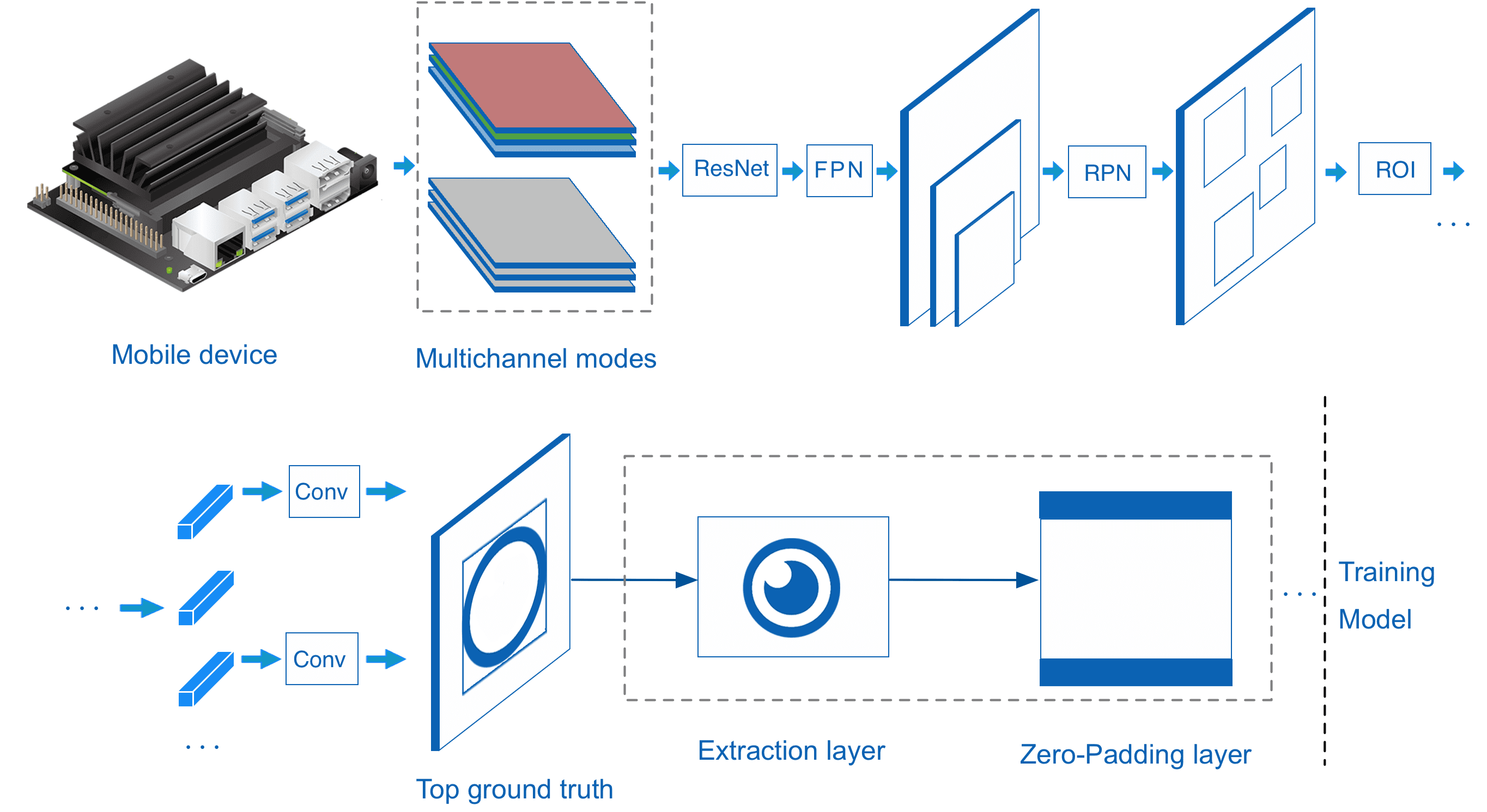}
\mbox{\bf }  
\begin{flushleft}{\textbf{Fig 7. The optimized Mask R-CNN architectures in the mobile environment.} The \textit{first} dotted box illustrates that we implement two input modes are applied during the preprocessing of the system, \textit{VW(Color)} and \textit{NIR(Grey)} modes. The \textit{second} dotted box illustrates that the iris features map with discriminative information was obtained by our custom \textit{extraction layer} and these scale-variant features are processed by the custom \textit{zero-padding layers}. Each individual component proposed in our Mask R-CNN architecture does influence iris recognition performance.}
\rule[0.25\baselineskip]{380pt}{0.3pt}
\end{flushleft}
\label{fig}
\end{figure}

There are two essential components in the TensorFlow, which are the convolution layer and the pooling layer. The convolution output sizes obtained by the two modes are always different \cite{ref49}. We can choose the “Valid” and “Same” modes to handle the padding mode of the output data. The purpose of padding component is to determine whether to fill the additional edge pixels of the input image matrix when performing convolution or pooling operations. Based on this concept, in order to enable the images processed by the proposed Mask R-CNN to be trained in the further recognition model, we have customized a zero-padding layer in the architecture of Mask R-CNN to make the shapes match of the output image as needed, and directly input into the further recognition model, as shown in Fig 7.

\begin{figure}[!h]
\centerline{\includegraphics[width=13.5cm]{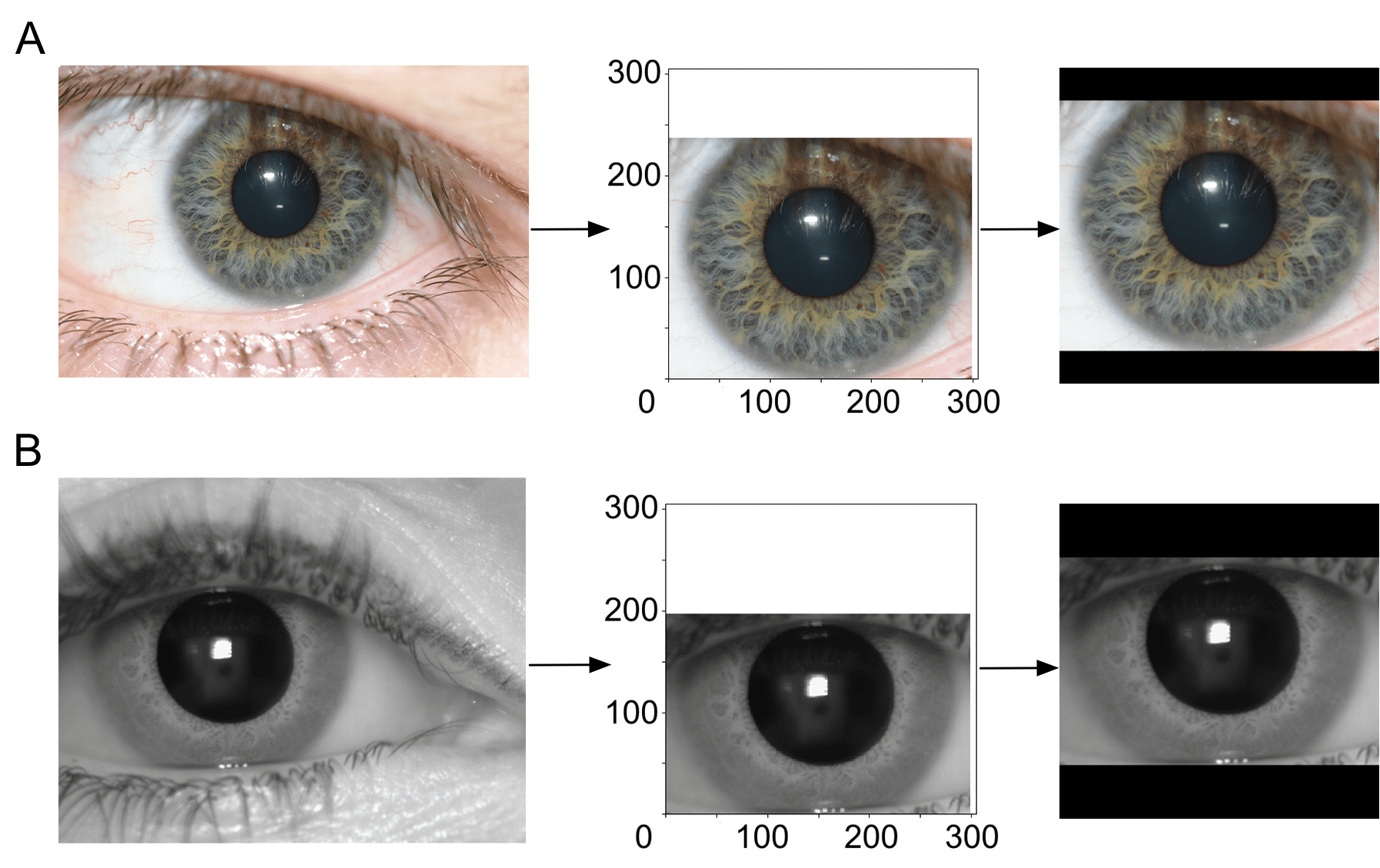}}

\begin{flushleft}{\textbf{Fig 8. The proposed zero-padding normalization on extracted iris images with different sizes.} The experimental results show that the iris region is extracted from the raw data (\textit{left}, original iris image), then the extracted scale-variant iris features (\textit{middle}, iris region) is transferred into a standard training set(\textit{right}, processed iris image with padding layer).}

\rule[0.25\baselineskip]{\textwidth}{0.3pt}
\end{flushleft}
\label{fig}
\end{figure}


Most importantly, our proposed framework can efficiently enhance the robustness of the model output. It can extract the iris region by the extraction layer, and the custom padding layer can fill the vacant area with pixels in the vertical direction and the width kept the same. All the value of filled pixels is 0 with black color \cite{ref49} \cite{ref50}. The top row represents the color iris image, while the bottom row represents the gray iris image. Finally, all iris images are cropped to the 299×299 pixels feature maps, as displayed in Fig 8. The advantage is that it can reduce GPU computation during the training model, and the stability of the recognition model for image processing is enhanced.

\quad
\subsection{The Fine-Tuned Mobile Inception V4 Architecture}
The fine-tuned mobile Inception V4 is executed to identify the human iris. For our experimental dataset, the iris image represents a specific domain, such as iris periocular and iris characters, which belongs to the human special central visual system. For these particular texture images, a strategic priority for us is that we would fine-tune the Inception V4 neural networks continue training it on the iris dataset we have.  
  
Many novel models and efficient learning techniques have been introduced to make CNN's model deeper and powerful \cite{ref51} \cite{ref52}, achieving revolutionary performance in a wide range of inputted data. Szegedy et al. \cite{ref53} proposed an improved mobile version of Inception v4 based on Inception v3 \cite{ref54}. Inception v4 primarily consists of an input stem, three different Inceptions and two Reduction modules. Based on this structure, we have developed a fine-tuned model based on Inception v4. The proposed model includes a final classifier layer, and its dimensions differ from the original model. Fig 9 shows the complete architecture of the proposed model and the part of the fine-tuning with the dashed box. The implement of iris recognition on a mobile device is significantly different from the environment on a dedicated device, the former relies on the computational power and the limited space for storage.

\begin{figure}[!h]
\centerline{\includegraphics[width=14cm]{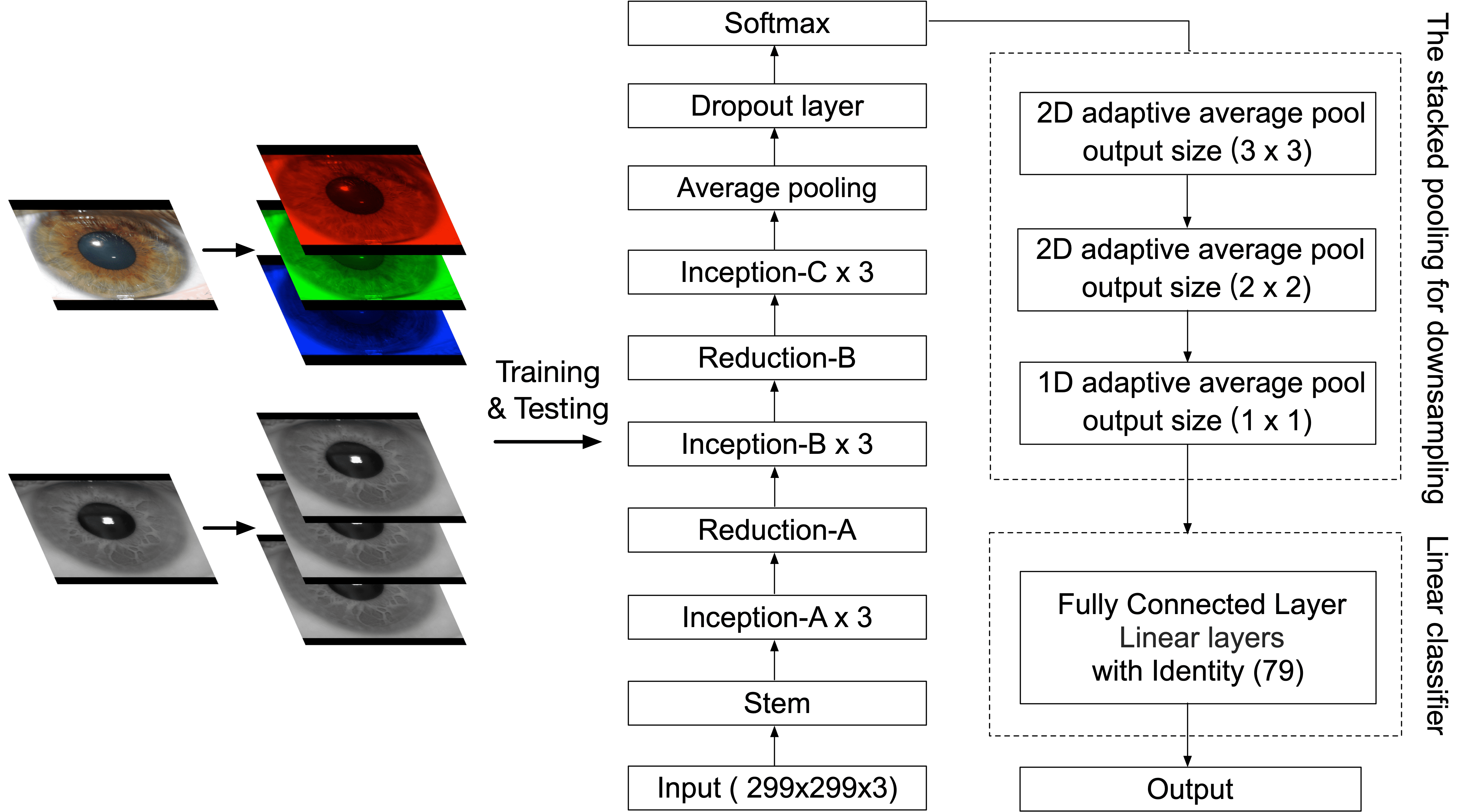}}

\mbox{\bf }  

\begin{flushleft}{\mbox{\bf Fig 9. The fine-tuned Inception V4 architecture.} The final pooling procedure is described as a series of adaptive pooling combinations, where the filter is with different output sizes. The final full connection layer is adjusted according to the number of classifications in this experiment. These two subcomponents (\textit{with two dotted boxes above}) are the optimized structures generated by our fine-tuning method.}

\rule[0.25\baselineskip]{\textwidth}{0.3pt}
\end{flushleft}
\label{fig}
\end{figure}

Below are the implementation datils of our fine-tuned Inception V4 neural network model:

First of all, we fine-tuned the first layer of our model based on the original Inception V4 model. The input size is adjusted to 299×299 pixels.  Different networks require different input sizes. For example, our fine-tuned Inception V4 requires that the image size of the input network is 299×299 pixels. Therefore, before the iris image is input to the proposed network, images need to be downsampled in the specified size for network input.  

To make the model recognize the iris more efficiently in a mobile environment, we loaded the pre-training weights and custom last classification layers to develop our model. Our task is to make the network focus on learning scale-variant features in the final linear layer. Siyu Huang et al. propose simple but effective variants of the pooling module, named stacked pooling \cite{ref55}. The stacked pooling is an equivalent form of multi-kernel pooling \cite{ref56} by stacking smaller kernel pooling. All the pooling operations are calculated on down-sampled feature maps except for its first kernel pooling, which can reduce the computing cost for training the learning model. Their empirical studies reveal that the stacked pooling shows a better computing efficiency than multi-kernel pooling. In our fine-tuned model, we stacked three adaptive average pooling layers, the first two layers are two-dimensional, and the output sizes are 3×3 and 2×2, respectively. The last one output size is 1 that applies a 1D adaptive average pooling over an input signal composed of several input planes. 

In addition, our iris recognition is a classification task on the UTiris datasets, the new fully connected layer of the proposed network will be of 79 categories instead of the original 1000 categories. The adjustment is to reconstruct the fully connected layer (the softmax layer), which is implemented by a linear layer combining with the identity operator. 

A good selection of activation function is an important part of the design in a neural network. In the process of backpropagation, the gradient has direction and size. The gradient descent algorithm multiplies a variable called the learning rate to determine the location of the next point. A paper \cite{ref57} by Leslie N. Smith describes some very instructive learning rate settings to find the initial learning rate. If the learning rate was low, the gradient was decline slowly, and the training took a long time. By contrast, if the learning rate setting was large, then it was difficult to converge to the extreme value. Considering the experiment in practice, we use the value of 1e-4 as the learning rate in our proposed model.  

Finally, we need to use an optimization algorithm to iterate over the model parameters to minimize the loss function value. In many fields such as computer vision, the most commonly used is the gradient descent method to find the optimal loss during the training phase \cite{ref58}. Although the Adam algorithm is currently the mainstream optimization algorithm, some researchers also pointed out the defect of Adam's convergence \cite{ref59} \cite{ref60}. Compared with Adam, the AMSGrad gradient descent function is more robust to the parameter changes and is more stable in the training process \cite{ref61}. Thus, AMSGrad can substitute for the regular Adam without the momentum parameter in our proposed model.

Our proposed model is to adjust and add the learning layers to optimize the Inception V4 and investigate whether they can improve decision making in the recognition process in the mobile environment. The methodologies above details the implementation of the multi-task deep learning architectures for single iris object detection on our edge computing device.

\quad
\subsection{The Iris Datasets} 
To evaluate the performance of the proposed framework, we selected one well-known iris dataset of UTiris, which includes the iris image acquisition scheme by using different devices \cite{ref44}. The database consists of a total of 1540 iris images, including the Visible-Wavelength (VW) and the Near-Infrared (NIR) session. All the iris images are captured under non-constraint conditions, such as non-ideal images, imaging distance, and illumination conditions. There are 804 images of 2048×1360 pixels in the VW session, and others 736 images of 1000×776 pixels in the NIR session, are taken from 79 individuals demonstrated in 158 classifications. The dataset is collected by the University of Tehran during 24-27th of June 2007.   

\quad
\subsection{The K-fold Cross Validation}
The cross-validation \cite{ref62} emphasizes the objective evaluation of the matching degree of the model to the data. This data analysis method \cite{ref63} divides the original data into k groups, makes a verification for each subset and takes the rest of the k-1 subset data as the training set, thus the cross-validation can train K models. Particularly, the K models are evaluated in the validation set, and the final loss is obtained by the weighted mean operation via the loss of each model. The loss function of our recognition model is implemented by the cross-entropy loss function, combining the log softmax and negative log likelihood functions to calculate the model loss. This strategy is capable of measuring subtle differences and is suitable for applying in fine-grained image classification tasks.

\quad
\subsection{The Edge Calculation Devices}  
The integrated framework above is executed on a GPU-based edge calculation device - Jetson Nano. The Jetson Nano is proposed by Nvidia \cite{ref64}, which is intended for low-power applications requiring high computational performance in mobile environments. It equipped with Maxwell GPU with Quad ARM Cortex-A57 processor, and 4GB of LPDDR4 memory. Learning models can run with Linux kernel 3.10.96 on the Ubuntu 18.04 system. Design of low-power consumption (5 watts) and integrated GPU makes the Jetson Nano become an ideal candidate for conducting the proposed methods in this research. Due to its strong computing capabilities \cite{ref65}, researchers can perform multiple neural networks by parallel mode in certain fields, such as image classification, segmentation, and object detection in the complex mobile environment.

\quad
\section{\rmfamily Results}
\subsection{The Evaluation of Proposed Mask R-CNN Architecture}
In our experiment, we randomly use some portion of the iris image as the training set and then use the Callback function to save the best weights and load them before predicting the new iris images. There are approximately 20\% data for training and 80\% for testing the proposed Mask R-CNN. When finished the preprocessing phase, all iris images were fixed to 299×299 pixel to by the normalization operation in order to meet the model input size, this can efficiently reduce the capacity of the training set and increase the training speed for the further training model, as shown in Fig 10.

\begin{figure}[!h]
\includegraphics[width=13.7cm]{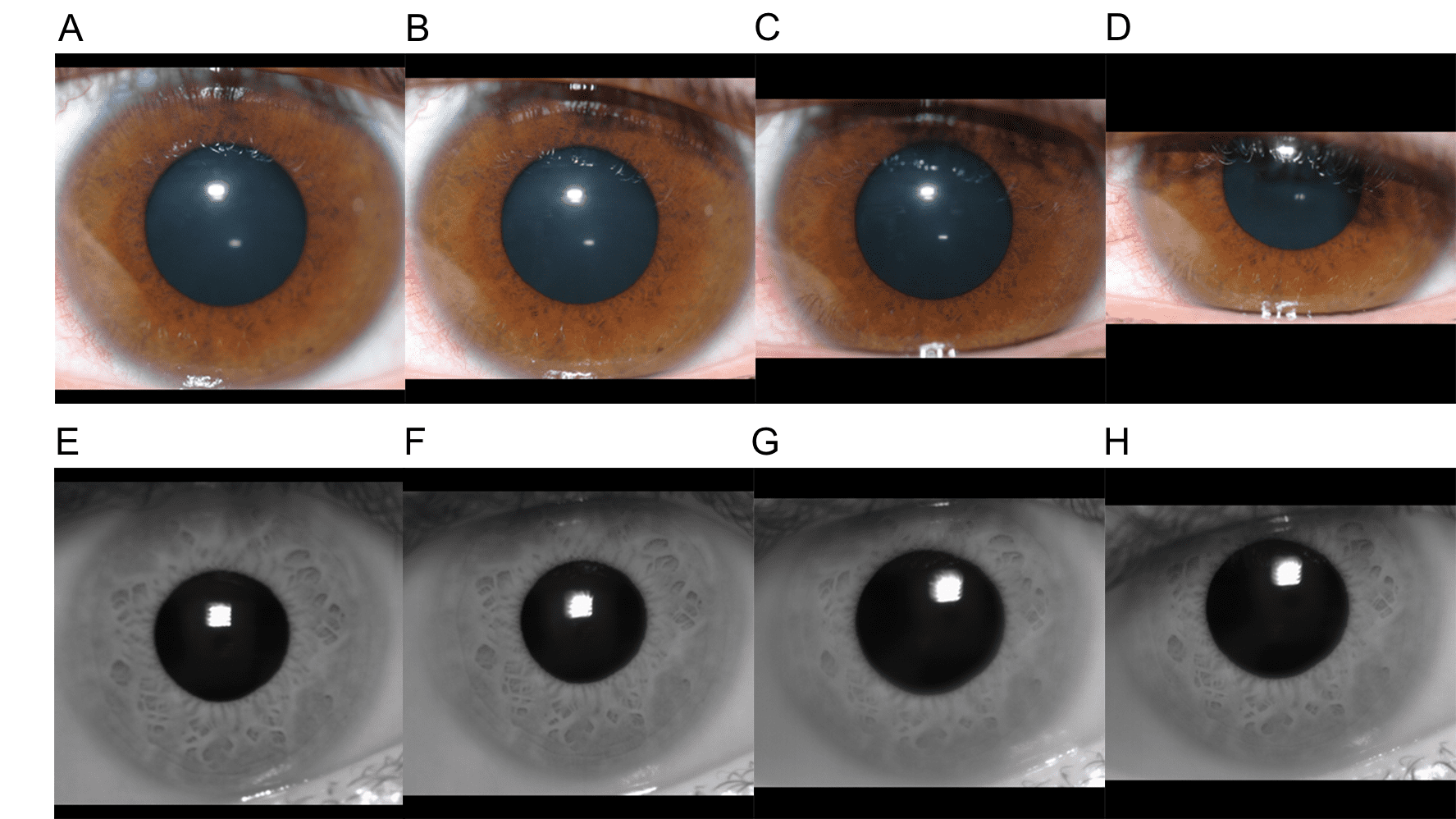}
\begin{flushleft}{\mbox{\bf  Fig 10.Different degrees of padding processing effects.} Comparing the different degrees of image processing effects with the color (Fig 10A-10D) and gray iris images (Fig 10E-10H). The \textit{left}-to-\textit{right} images represent the padding effect sorted the iris by size, from smallest to largest in padding area.} 

\rule[0.25\baselineskip]{\textwidth}{0.3pt}
\end{flushleft}
\label{fig}
\end{figure}

The experiment results in Fig 10 show that the effect of experimental images, which testified the effectiveness and practicality of the proposed Mask R-CNN architecture. Different noise factors can be distinguished from the Fig 10 such as multiple scales, color distortion, and insufficient light. Among the most important is that some eyes are open while the other eye half-open or closed, which affects the performance and stability in the biometric identification process. Adding the zero-padding layer in the architecture is the implementation of the solution to our research problems, which can efficiently learn the scale-variant features from each iris training set and enhance the robustness of the iris authentication system. The proposed Mask R-CNN also uses a multi-task loss function to calculate the loss, combining the loss of classification, localization and segmentation mask. The equation is defined as below. 
\begin{align*}
Loss &= L_{cls}+L_{box}+L_{mask} 
\end{align*}

The formula consists of Loss functions in each ROI region: classification loss and position regression loss of the bounding box. They were inherited from Faster R-CNN and the Loss of the mask is proposed by \cite{ref42}. During training the model, the Loss evidently dropped to nearly 0.01\% and to the lowest of 0.0066\% at 203rd and 233rd of 250 epochs, respectively. Our proposed Mask R-CNN architecture correctly identified 1539 of 1540 iris images in UTiris datasets, the fail one is due to the improper capture of partial iris images. Finally, the precision of iris detection is more than 99.99\%. This section summarizes our Mask R-CNN architecture, including the stronger robustness of detecting the iris location and the extracting the precise region of the iris region, are presented. A baseline dataset has been completed that serves as a basis to measure evaluation targets. Next, we comprehensively validate the performance of the proposed framework by using a pre-processed iris dataset.


\quad
\subsection{The Evaluation of Fine-tuned Mobile Inception V4 Architecture}

In the recognition phase, our framework was conducted on UTiris datasets and analyze the performance of the fine-tuned mobile Inception V4. The total size of the training set is 1540 images, which were randomly sorted in the memory. The learning rate of the whole network was initialized to 1e-4. The testing iterations was set to 17 epochs on the mini-batches of eight images during the training, using the AMSGrad optimization algorithm. All of these super-parameters were set properly so that the network be able to generalize well. Simultaneously, these factors could be used to improve the performance and accuracy of our iris authentication systems.  

\begin{figure}[!h]
\centerline{\includegraphics[width=13.5cm]{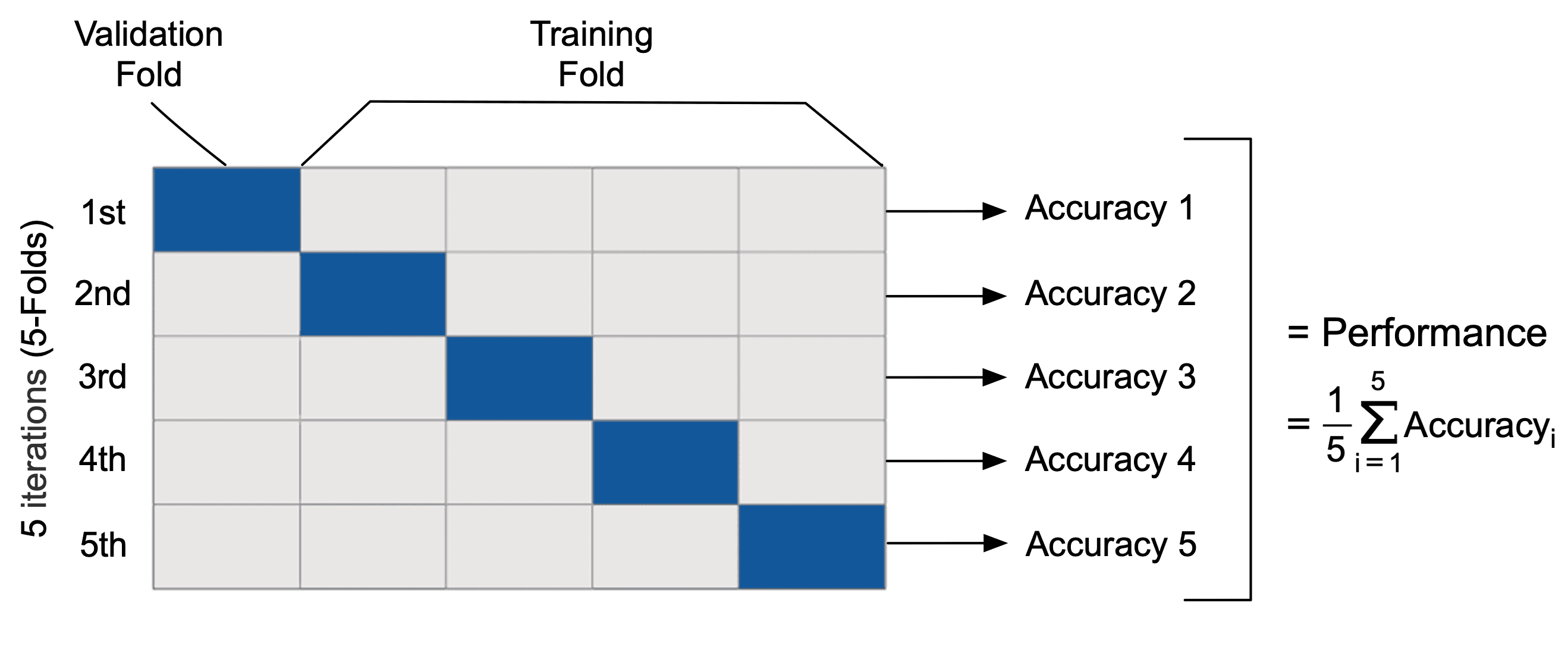}}
\begin{flushleft}{\mbox{\bf \centerline{ Fig 11. The 5-fold cross validation}}}

\end{flushleft}
\label{fig}
\end{figure}

The cross-validation strategy was employed to verify the performance. As illustrated in Fig 11, we applied \textit{Five fold cross validation} \cite{ref66} to evaluate the performance of our proposed mobile Inception V4 on the Nvidia Jetson Nano. Every sub dataset of UTiris we evaluated, and the accuracy rate verified the measurement of the recognition learning model.

\begin{figure}[!h]
\includegraphics[width=14cm]{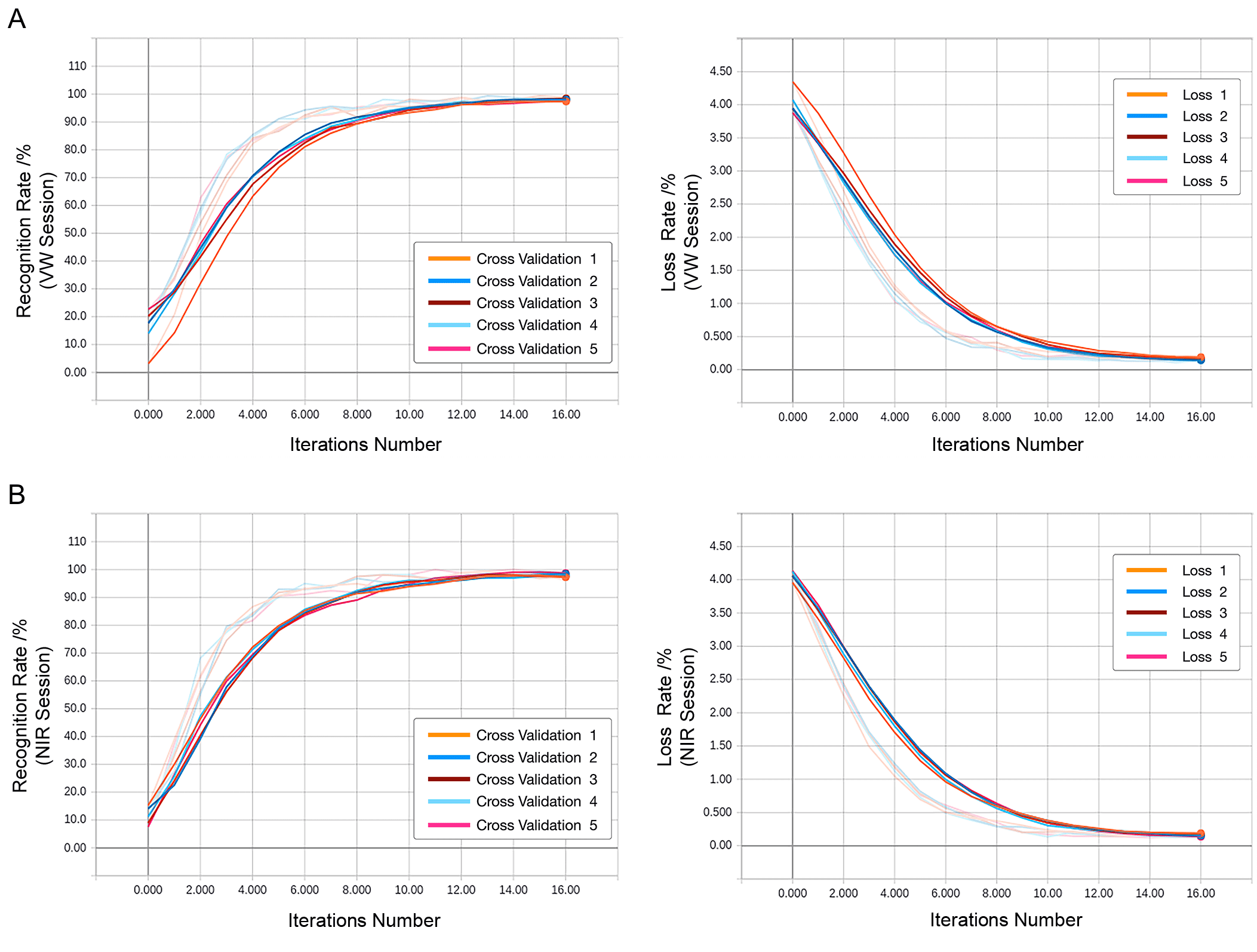}
\begin{flushleft}{\textbf{Fig 12. The curve plot for the experimental result of 5-fold cross validation.} The X-axis displays the number of epochs ranging from \textit{0} to \textit{16}, and the Y-axis displays the validation accuracy from \textit{0} to \textit{110}. Curve plot for the categorical accuracy for \textit{17} epochs. The accuracy was defined as the total frequency of the correct acceptance and the imitators with correct rejection.} 
\\
\rule[0.25\baselineskip]{390pt}{0.3pt}
\end{flushleft}
\label{fig}
\end{figure}

The final curve of validation accuracy and loss were used to evaluate the proposed framework, as displayed in Fig 12. Through the training of the proposed learning model with 17 epochs, all experimental results are visualized by the Tensorboard tool. The figures of Fig 12(A) and Fig 12(C) illustrated the accuracy is proportional to the increasing epochs and remains steady accuracy after the 12th epoch. In each cross-validation, we chose the highest accuracy as the benchmark to represent the best performance of our fine-tuned learning model. In particular, the experimental results are presented in Fig 12(A), it manifests that while mobile inception V4 is applied on VW session, the best result was achieved by the fine-tuned method with the accuracy of 99.37\% and peaked at 100\% recognition rate in 5th cross-validation of 11th epoch on NIR session. Here, the difference between the average accuracy of the proposed framework with VW and NIR was 0.3\%. Gradually, with the help of optimization functions, loss function learned to reduce the error in prediction. The results showed that the average loss rate is decreased by roughly 4\%, from 4.05\% to 0.14\%. We found that increasing the number of training iterations could significantly improve the proposed learning framework performance. Comparative analysis indicates the experimental results could be acceptable for testing models that could be as close as possible to the performance; this is because the cross-validation takes effectively use of the generated data.

\quad
\section{\rmfamily Discussion}
In this study, an iris authentication system based on Nvidia Jetson Nano was introduced. The proposed system consists of two distinct functions, iris detection and recognition. We consider the iris database, which consists of 2 classes. The dataset is split into training data and testing data set and organized into 2 folders, train and test. The 5- fold cross-validations are calculated with a batch size of 8 for 17 epochs, demonstrating that the fine-tuned mobile Inception V4 model was steady in every validation phase, as explained in Tables 1 and 2.




\begin{table}[!ht]
\begin{adjustwidth}{-0.5in}{0in}


\textbf{\footnotesize Table 1: The table presents the intercorrelations among the 5-fold cross- validations on VW session.}

\rowcolors {1}{white}{mygray}
\tiny
\setlength{\tabcolsep}{1mm}
\begin{tabular}{p{1.4cm}|p{0.7cm}|p{0.7cm}|p{0.7cm}|p{0.7cm}|p{0.7cm}|p{0.7cm}|p{0.7cm}|p{0.7cm}|p{0.7cm}|p{0.7cm}|p{0.7cm}|p{0.7cm}|p{0.7cm}|p{0.7cm}|p{0.7cm}|p{0.7cm}|p{0.7cm}p{0.0cm}}
 

\hline
\tabincell{c}{Fine-tuned\\Inception V4}& \tabincell{c}{Epoch\\0} &\tabincell{c}{Epoch\\1} &\tabincell{c}{Epoch\\2} &\tabincell{c}{Epoch\\3} &\tabincell{c}{Epoch\\4} &\tabincell{c}{Epoch\\5} & \tabincell{c}{Epoch\\6} & \tabincell{c}{Epoch\\7} &\tabincell{c}{Epoch\\8} &\tabincell{c}{Epoch\\9}& \tabincell{c}{Epoch\\10} &\tabincell{c}{Epoch\\11} &\tabincell{c}{Epoch\\12} &\tabincell{c}{Epoch\\13} & \tabincell{c}{Epoch\\14} & \tabincell{c}{Epoch\\15} & \tabincell{c}{Epoch\\16} &  \\ \hline
\tabincell{c}{Cross\\ Validation \\ 1}& \tabincell{c}{3.163} &\tabincell{c}{20.89} &\tabincell{c}{49.37} &\tabincell{c}{68.35} &\tabincell{c}{82.28} &\tabincell{c}{87.97} & \tabincell{c}{91.77} & \tabincell{c}{93.04} &\tabincell{c}{94.30} &\tabincell{c}{95.57}& \tabincell{c}{95.57} &\tabincell{c}{96.20} &\tabincell{c}{98.71} &\tabincell{c}{96.84} & \tabincell{c}{98.10} & \tabincell{c}{97.47} & \tabincell{c}{97.47} &  \\ \hline
\tabincell{c}{Validation \\ Loss}& \tabincell{c}{4.347} &\tabincell{c}{3.584} &\tabincell{c}{2.691} &\tabincell{c}{1.865} &\tabincell{c}{1.247} &\tabincell{c}{0.8529} & \tabincell{c}{0.5879} & \tabincell{c}{0.4328} &\tabincell{c}{0.3355} &\tabincell{c}{0.3285}& \tabincell{c}{0.2723} &\tabincell{c}{0.2568} &\tabincell{c}{0.1840} &\tabincell{c}{0.2100} & \tabincell{c}{0.1538} & \tabincell{c}{0.1724} & \tabincell{c}{0.1787} &  \\ \hline

\tabincell{c}{Cross\\ Validation \\ 2}& \tabincell{c}{17.72} &\tabincell{c}{37.34} &\tabincell{c}{58.86} &\tabincell{c}{76.58} &\tabincell{c}{85.44} &\tabincell{c}{91.14} & \tabincell{c}{94.30} & \tabincell{c}{95.57} &\tabincell{c}{94.94} &\tabincell{c}{95.57}& \tabincell{c}{97.47} &\tabincell{c}{97.47} &\tabincell{c}{97.47} &\tabincell{c}{98.37} & \tabincell{c}{98.73} & \tabincell{c}{98.10} & \tabincell{c}{98.10} &  \\ \hline
\tabincell{c}{Validation \\ Loss}& \tabincell{c}{3.947} &\tabincell{c}{3.095} &\tabincell{c}{2.354} &\tabincell{c}{1.652} &\tabincell{c}{1.146} &\tabincell{c}{0.7699} & \tabincell{c}{0.4740} & \tabincell{c}{0.3398} &\tabincell{c}{0.3281} &\tabincell{c}{0.2530}& \tabincell{c}{0.1713} &\tabincell{c}{0.1869} &\tabincell{c}{0.1559} &\tabincell{c}{0.1331} & \tabincell{c}{0.1291} & \tabincell{c}{0.1373} & \tabincell{c}{0.1269} &  \\ \hline

\tabincell{c}{Cross\\ Validation \\ 3}& \tabincell{c}{20.25} &\tabincell{c}{34.18} &\tabincell{c}{53.80} &\tabincell{c}{70.89} &\tabincell{c}{84.18} &\tabincell{c}{86.71} & \tabincell{c}{92.41} & \tabincell{c}{95.57} &\tabincell{c}{91.77} &\tabincell{c}{94.94}& \tabincell{c}{98.10} &\tabincell{c}{97.47} &\tabincell{c}{98.73} &\tabincell{c}{97.47} & \tabincell{c}{98.10} & \tabincell{c}{99.37} & \tabincell{c}{98.73} &  \\ \hline
\tabincell{c}{Validation \\ Loss}& \tabincell{c}{3.940} &\tabincell{c}{3.164} &\tabincell{c}{2.489} &\tabincell{c}{1.753} &\tabincell{c}{1.217} &\tabincell{c}{0.8743} & \tabincell{c}{0.5731} & \tabincell{c}{0.4074} &\tabincell{c}{0.4083} &\tabincell{c}{0.2827}& \tabincell{c}{0.1922} &\tabincell{c}{0.1877} &\tabincell{c}{0.1621} &\tabincell{c}{0.1836} & \tabincell{c}{0.1559} & \tabincell{c}{0.1482} & \tabincell{c}{0.1365} &  \\ \hline

\tabincell{c}{Cross\\ Validation \\ 4}& \tabincell{c}{14.01} &\tabincell{c}{36.93} &\tabincell{c}{57.32} &\tabincell{c}{78.34} &\tabincell{c}{84.71} &\tabincell{c}{91.08} & \tabincell{c}{91.08} & \tabincell{c}{94.90} &\tabincell{c}{94.27} &\tabincell{c}{98.09}& \tabincell{c}{97.45} &\tabincell{c}{97.45} &\tabincell{c}{98.73} &\tabincell{c}{96.82} & \tabincell{c}{97.45} & \tabincell{c}{98.73} & \tabincell{c}{98.73} &  \\ \hline
\tabincell{c}{Validation \\ Loss}& \tabincell{c}{4.073} &\tabincell{c}{3.067} &\tabincell{c}{2.219} &\tabincell{c}{1.596} &\tabincell{c}{1.054} &\tabincell{c}{0.7190} & \tabincell{c}{0.5629} & \tabincell{c}{0.3877} &\tabincell{c}{0.3307} &\tabincell{c}{0.1651}& \tabincell{c}{0.1555} &\tabincell{c}{0.1583} &\tabincell{c}{0.1314} &\tabincell{c}{0.1817} & \tabincell{c}{0.1509} & \tabincell{c}{0.1040} & \tabincell{c}{0.1358} &  \\ \hline

\tabincell{c}{Cross\\ Validation \\ 5}& \tabincell{c}{22.67} &\tabincell{c}{33.33} &\tabincell{c}{62.67} &\tabincell{c}{77.33} &\tabincell{c}{83.33} &\tabincell{c}{87.33} & \tabincell{c}{92.00} & \tabincell{c}{92.67} &\tabincell{c}{95.33} &\tabincell{c}{96.00}& \tabincell{c}{97.33} &\tabincell{c}{96.00} &\tabincell{c}{98.67} &\tabincell{c}{96.00} & \tabincell{c}{97.33} & \tabincell{c}{98.00} & \tabincell{c}{98.00} &  \\ \hline
\tabincell{c}{Validation \\ Loss}& \tabincell{c}{3.879} &\tabincell{c}{3.133} &\tabincell{c}{2.200} &\tabincell{c}{1.593} &\tabincell{c}{1.031} &\tabincell{c}{0.7751} & \tabincell{c}{0.5743} & \tabincell{c}{0.4867} &\tabincell{c}{0.2955} &\tabincell{c}{0.2077}& \tabincell{c}{0.1971} &\tabincell{c}{0.2256} &\tabincell{c}{0.1518} &\tabincell{c}{0.1867} & \tabincell{c}{0.1798} & \tabincell{c}{0.1435} & \tabincell{c}{0.1333} &  \\ \hline

\end{tabular}
\\
\begin{flushleft} \small Table notes that all the accuracy rates/\% and loss rates/\% of 17 epochs measures on the proposed framework in columns, and the rows show five iterations for each k cross-validation(k=5).

\rule[0.25\baselineskip]{492pt}{0.3pt}
\end{flushleft}
\label{table1}

\end{adjustwidth}
\end{table}


\begin{table}[!ht]

\begin{adjustwidth}{-0.5in}{0in}


\textbf{\footnotesize Table 2: The table presents the intercorrelations among the 5-fold cross- validations on NIR session.}

\rowcolors {1}{white}{mygray}
\tiny
\setlength{\tabcolsep}{1mm}
\begin{tabular}{p{1.4cm}|p{0.7cm}|p{0.7cm}|p{0.7cm}|p{0.7cm}|p{0.7cm}|p{0.7cm}|p{0.7cm}|p{0.7cm}|p{0.7cm}|p{0.7cm}|p{0.7cm}|p{0.7cm}|p{0.7cm}|p{0.7cm}|p{0.7cm}|p{0.7cm}|p{0.7cm}p{0.0cm}}
 

\hline
\tabincell{c}{Fine-tuned\\Mobile Inception V4}& \tabincell{c}{Epoch\\0} &\tabincell{c}{Epoch\\1} &\tabincell{c}{Epoch\\2} &\tabincell{c}{Epoch\\3} &\tabincell{c}{Epoch\\4} &\tabincell{c}{Epoch\\5} & \tabincell{c}{Epoch\\6} & \tabincell{c}{Epoch\\7} &\tabincell{c}{Epoch\\8} &\tabincell{c}{Epoch\\9}& \tabincell{c}{Epoch\\10} &\tabincell{c}{Epoch\\11} &\tabincell{c}{Epoch\\12} &\tabincell{c}{Epoch\\13} & \tabincell{c}{Epoch\\14} & \tabincell{c}{Epoch\\15} & \tabincell{c}{Epoch\\16} &  \\ \hline
\tabincell{c}{Cross\\ Validation \\ 1}& \tabincell{c}{15.29} &\tabincell{c}{38.85} &\tabincell{c}{61.78} &\tabincell{c}{78.34} &\tabincell{c}{86.62} &\tabincell{c}{90.45} & \tabincell{c}{92.99} & \tabincell{c}{94.27} &\tabincell{c}{94.90} &\tabincell{c}{93.63}& \tabincell{c}{96.18} &\tabincell{c}{96.18} &\tabincell{c}{98.73} &\tabincell{c}{99.36} & \tabincell{c}{98.09} & \tabincell{c}{98.09} & \tabincell{c}{96.18} &  \\ \hline
\tabincell{c}{Validation \\ Loss}& \tabincell{c}{3.958} &\tabincell{c}{3.083} &\tabincell{c}{2.254} &\tabincell{c}{1.488} &\tabincell{c}{1.046} &\tabincell{c}{0.6934} & \tabincell{c}{0.5036} & \tabincell{c}{0.4207} &\tabincell{c}{0.3747} &\tabincell{c}{0.3080}& \tabincell{c}{0.2304} &\tabincell{c}{0.2041} &\tabincell{c}{0.1837} &\tabincell{c}{0.1500} & \tabincell{c}{0.1783} & \tabincell{c}{0.1879} & \tabincell{c}{0.1806} &  \\ \hline

\tabincell{c}{Cross\\ Validation \\ 2}& \tabincell{c}{14.10} &\tabincell{c}{27.56} &\tabincell{c}{55.13} &\tabincell{c}{79.49} &\tabincell{c}{83.33} &\tabincell{c}{92.95} & \tabincell{c}{92.95} & \tabincell{c}{93.95} &\tabincell{c}{96.79} &\tabincell{c}{95.51}& \tabincell{c}{96.15} &\tabincell{c}{96.79} &\tabincell{c}{97.44} &\tabincell{c}{98.72} & \tabincell{c}{98.08} & \tabincell{c}{99.36} & \tabincell{c}{98.72} &  \\ \hline
\tabincell{c}{Validation \\ Loss}& \tabincell{c}{4.061} &\tabincell{c}{3.282} &\tabincell{c}{2.411} &\tabincell{c}{1.710} &\tabincell{c}{1.236} &\tabincell{c}{0.8172} & \tabincell{c}{0.5813} & \tabincell{c}{0.4011} &\tabincell{c}{0.2987} &\tabincell{c}{0.2820}& \tabincell{c}{0.2378} &\tabincell{c}{0.1842} &\tabincell{c}{0.1670} &\tabincell{c}{0.1650} & \tabincell{c}{0.1443} & \tabincell{c}{0.1249} & \tabincell{c}{0.1407} &  \\ \hline

\tabincell{c}{Cross\\ Validation \\ 3}& \tabincell{c}{8.91} &\tabincell{c}{33.12} &\tabincell{c}{56.05} &\tabincell{c}{74.52} &\tabincell{c}{84.08} &\tabincell{c}{91.72} & \tabincell{c}{92.99} & \tabincell{c}{93.63} &\tabincell{c}{97.45} &\tabincell{c}{98.09}& \tabincell{c}{97.45} &\tabincell{c}{96.82} &\tabincell{c}{98.73} &\tabincell{c}{99.36} & \tabincell{c}{98.09} & \tabincell{c}{96.82} & \tabincell{c}{97.45} &  \\ \hline
\tabincell{c}{Validation \\ Loss}& \tabincell{c}{4.061} &\tabincell{c}{3.237} &\tabincell{c}{2.437} &\tabincell{c}{1.695} &\tabincell{c}{1.179} &\tabincell{c}{0.7826} & \tabincell{c}{0.5643} & \tabincell{c}{0.4510} &\tabincell{c}{0.2988} &\tabincell{c}{0.2027}& \tabincell{c}{0.2073} &\tabincell{c}{0.2127} &\tabincell{c}{0.1461} &\tabincell{c}{0.1474} & \tabincell{c}{0.1594} & \tabincell{c}{0.1854} & \tabincell{c}{0.1444} &  \\ \hline

\tabincell{c}{Cross\\ Validation \\ 4}& \tabincell{c}{11.25} &\tabincell{c}{34.38} &\tabincell{c}{68.13} &\tabincell{c}{77.50} &\tabincell{c}{84.38} &\tabincell{c}{90.00} & \tabincell{c}{95.00} & \tabincell{c}{93.75} &\tabincell{c}{97.50} &\tabincell{c}{98.13}& \tabincell{c}{98.13} &\tabincell{c}{95.63} &\tabincell{c}{98.75} &\tabincell{c}{96.88} & \tabincell{c}{96.88} & \tabincell{c}{98.75} & \tabincell{c}{98.13} &  \\ \hline
\tabincell{c}{Validation \\ Loss}& \tabincell{c}{4.124} &\tabincell{c}{3.183} &\tabincell{c}{2.256} &\tabincell{c}{1.657} &\tabincell{c}{1.125} &\tabincell{c}{0.7234} & \tabincell{c}{0.4915} & \tabincell{c}{0.3818} &\tabincell{c}{0.2901} &\tabincell{c}{0.2054}& \tabincell{c}{0.1326} &\tabincell{c}{0.2061} &\tabincell{c}{0.1663} &\tabincell{c}{0.2056} & \tabincell{c}{0.1669} & \tabincell{c}{0.1421} & \tabincell{c}{0.1258} &  \\ \hline

\tabincell{c}{Cross\\ Validation \\ 5}& \tabincell{c}{7.59} &\tabincell{c}{37.34} &\tabincell{c}{61.39} &\tabincell{c}{78.48} &\tabincell{c}{81.65} &\tabincell{c}{95.51} & \tabincell{c}{91.14} & \tabincell{c}{92.41} &\tabincell{c}{91.77} &\tabincell{c}{98.10}& \tabincell{c}{98.10} &\tabincell{c}{100} &\tabincell{c}{98.73} &\tabincell{c}{99.37} & \tabincell{c}{100} & \tabincell{c}{99.37} & \tabincell{c}{98.10} &  \\ \hline
\tabincell{c}{Validation \\ Loss}& \tabincell{c}{4.134} &\tabincell{c}{3.319} &\tabincell{c}{2.374} &\tabincell{c}{1.681} &\tabincell{c}{1.192} &\tabincell{c}{0.7588} & \tabincell{c}{0.6120} & \tabincell{c}{0.4739} &\tabincell{c}{0.3495} &\tabincell{c}{0.2026}& \tabincell{c}{0.1707} &\tabincell{c}{0.1466} &\tabincell{c}{0.1456} &\tabincell{c}{0.1354} & \tabincell{c}{0.1218} & \tabincell{c}{0.1355} & \tabincell{c}{0.1128} &  \\ \hline

\end{tabular}
\\
\begin{flushleft}\small Table notes that all the accuracy rates/\% and loss rates/\% of 17 epochs measures on the proposed framework in columns, and the rows show five iterations for each k cross-validation(k=5).

\rule[0.25\baselineskip]{492pt}{0.3pt}
\end{flushleft}
\label{table1}
\end{adjustwidth}
\end{table}


In the linear layer of the fine-tuned mobile Inception V4 architecture, we propose a module of three adaptive average pooling layers with various properties and sizes. This module emphasizes the down sampling of the overall feature information, mainly include two positive roles: 1) while reducing the parameters, it is more reflected in the complete transmission of information. 2) more useful discriminative information is passed to the next layer for the feature extraction while reducing the dimension. These changes effectively speed up the fitting of depth network and data, and it decreases considerable computing costs so that our fine-tuned model can achieve successive and ideal recognition rates in the first few epochs. Fig 9 illustrates an example of the stacked pooling, where the kernel set K =\{3, 2, 1\} and the stride s = 1. In empirical studies, this configuration shows the best performance in most cases. Although these stacked layers increase the computational complexity of the proposed model, the research still demonstrates that the complex combination of pooling layers can significantly improve the recognition rate of the learning model more quickly in the neural network. Collectively, the proposed Mask R-CNN model has delivered a venerable of 99.99\% detection rate.

Considering the results in Tables 1 and 2, we can observe that increasing the number of training epochs can slightly improve the recognition rate. In the initial training phase, the recognition accuracy of the model in both sessions is relatively low. This is because the model needs to learn from the training data, whereas the sample of training data has not fitted the proposed model. On the whole, the average accuracy of VW sessions (15.56\%) is higher than the NIR session (11.43\%) in the first epoch since the color images contain a large amount of different discrete information in three channels. In the 8th epoch, the curve shows that the performance of the model on the NIR session (93.01\%) is smoothly beyond than the VW session (92.31\%). Based on the comparative analysis of two models, this research reveals the fact that the convergence speed of our fine-tined Inception V4 processing on the NIR session, which is better than the VW session in overall epochs. 

At the same time, we observe that the validation accuracy would fluctuate after 10th epochs. The reason for this is that the amount of data in the experimental data set is still relatively small, and the training on the model with more layers may bring some problems like overfitting due to the various size of pre-processed iris images and the effective recognition area within the range of each pre-processed iris image is disparate. Under the mobile environment, this seems inevitable. To solve this problem, we used the early stopping method that calculated the accuracy of validation data at the end of each epoch and stopped training when the accuracy is no longer improved. 

We also investigate how to train a fine-tuned CNN that can classify iris images with high accuracy. The hyperparameters of the learning rate are closely associated with model performance. Weight decay can be effective at preventing the problem of over-fitting. They control the size of network parameters updated after each iteration. Our experiment shows that when the learning rate is modulated by 1e-4, the proposed framework has a notable recognition rate of 100\% in the 5th verification set of VW session. Comparatively, the accuracy rate of assessment is over 99\% in NIR session. Likewise, we also try to use the learning rate of 1e-3 and 1e-5, whereas this leads to opposing effects of slow training or early occurrence of overfitting phenomenon because of an inappropriate learning rate and the insufficient training data \cite{ref67}.  

Lastly, compared to preprocessing and recognition approaches used by authors \cite{ref16} \cite{ref18} \cite{ref19} in Table 3, we make the framework have more agility and flexible space to adapt the change of system, and research demonstrates the effectiveness of the proposed framework. Indeed, degradation in image quality negatively impacts the performance of iris authentication systems, especially for the infrared images. Starting with good quality images, we do not adopt any degradation operation in our preprocessing phase. Our Mask R-CNN architecture is reconstructed to implement the robust detection and the scale-variant feature extraction, respectively. The model has the advantage that image space is unaffected by the image degradation process. Furthermore, our proposed recognition model of fine-tuned mobile Inception V4 always gets top performance for all levels of methodologies, demonstrating the robust adaptation and excellent performance and in iris detection, and the highest performance for iris recognition. It has been observed from the results that our framework provides the best average recognition accuracy of 98.97\% and 99.24\% for the VW and NIR session, respectively. The overall accuracy of the model in the UTiris dataset is 99.10\%, are summarized as follows.




\begin{table}[!ht]

\begin{adjustwidth}{-0.1in}{0in}
\small

\textbf{\footnotesize Table 3: Comparing the original results obtained from experiments with the state-of-the-art methodologies. The best results in bold.}

\rowcolors {1}{white}{white}
\footnotesize
\setlength{\tabcolsep}{1mm}
\begin{tabular}{p{2.5cm}|p{2.9cm}|p{1.9cm}|p{2cm}|p{3.5cm}p{0.0cm}}

\toprule[0.5pt]

\hline
\tabincell{c}{\quad References}& \tabincell{c}{\quad Preprocessing} &\tabincell{c}{Localization\\\ Precision} &\tabincell{c}{ Methodology} &\tabincell{c}{\quad  \quad \quad Recognition\\\quad \quad \quad Accuracy /\%} &  \\ \hline
\tabincell{c}{\ Bhagyashree\\ \ Deshpande\\ \ and
\ Deepak\\ \ Jayaswal \cite{ref16} }& \tabincell{c}{Daugman’s\\integro-differential\\+\\
Daugman’s\\Rubber Sheet\\+\\1D log Gabor filter} &\tabincell{c}{\quad\quad ×} &\tabincell{c}{ \   Hamming\\ \  Distance} &\tabincell{c}{ \quad \quad \quad \textbf{95.00\%} (VW)}&  \\ \hline
\tabincell{c}{\quad Mohammed\\ \quad Hamzah\\ \quad Abed \cite{ref18} }& \tabincell{c}{\quad Circular Hough\\\quad transform\\ \ + \\ Haar wavelet\\\quad transform\\ \ + \\\quad PCA} &\tabincell{c}{\quad 98.73\%} &\tabincell{c}{\ \ Cosine\\ \ \ Distance} &\tabincell{c}{\quad \quad \quad \textbf{91.14\%} (NIR)}&  \\ \hline

\tabincell{c}{Onkar Kaudki\\ and Kishor \\ Bhurchandi \cite{ref19} }& \tabincell{c}{\quad Rubber-Sheet\\\quad Unwrapping\\ \ \quad +\\ \quad Haar wavelet\\ \quad transform} &\tabincell{c}{\quad 96.21\%} &\tabincell{c}{\ \ Hamming\\ \ \ Distance} &\tabincell{c}{ \quad \quad \quad \textbf{97.00\%} (NIR)}&  \\ \hline

\tabincell{c}{\quad Proposed \\ \quad Frameworks }& \tabincell{c}{\ \quad  Custom \\ \ \quad Mask-RCNN} &\tabincell{c}{\quad  99.99\%} &\tabincell{c}{   Fine-tuned\\   Inception V4} &\tabincell{c}{ \ \quad \quad \textbf{ Accuracy/\%}}&  \\ \hline

\rule{0pt}{10pt} 
\tabincell{c}{}& \tabincell{c}{} &\tabincell{c}{} &\tabincell{c}{ \centerline{Session} } &\tabincell{c}{\quad \quad VW \ \quad \Big|  \  \quad NIR}&  \\ \hline

\rule{0pt}{10pt} 
\tabincell{c}{}& \tabincell{c}{} &\tabincell{c}{} &\tabincell{c}{ \centerline{CV 1} } &\tabincell{c}{\ \quad \textbf{98.71\%}  \quad \ \textbf{98.73\%}}&  \\ \hline

\rule{0pt}{10pt} 
\tabincell{c}{}& \tabincell{c}{} &\tabincell{c}{} &\tabincell{c}{ \centerline{CV 2} } &\tabincell{c}{\ \quad \textbf{99.37\%}  \quad \ \textbf{99.36\%}}&  \\ \hline

\rule{0pt}{10pt} 
\tabincell{c}{}& \tabincell{c}{} &\tabincell{c}{} &\tabincell{c}{ \centerline{CV 3} } &\tabincell{c}{\ \quad \textbf{99.37\%}  \quad \ \textbf{99.36\%}}&  \\ \hline

\rule{0pt}{10pt} 
\tabincell{c}{}& \tabincell{c}{} &\tabincell{c}{} &\tabincell{c}{ \centerline{CV 4} } &\tabincell{c}{\ \quad \textbf{98.73\%}  \quad \ \textbf{98.75\%}}&  \\ \hline

\rule{0pt}{10pt} 
\tabincell{c}{}& \tabincell{c}{} &\tabincell{c}{} &\tabincell{c}{ \centerline{CV 5} } &\tabincell{c}{\ \quad \textbf{98.67\%}  \quad \quad \textbf{100\%}}&  \\ \hline

\rule{0pt}{10pt} 
\tabincell{c}{}& \tabincell{c}{} &\tabincell{c}{} &\tabincell{c}{ \quad Average\\ \quad accuracy } &\tabincell{c}{\ \quad \textbf{98.97\%}  \quad \ \textbf{99.24\%}}&  \\ \hline

\rule{0pt}{10pt} 
\tabincell{c}{}& \tabincell{c}{} &\tabincell{c}{} &\tabincell{c}{ \quad Overall\\ \quad accuracy } &\tabincell{c}{\centerline{\textbf{\quad 99.10\%} }}&  \\ \hline

\bottomrule[0.5pt]
\end{tabular}
\\


\label{table1}
\end{adjustwidth}
\end{table}


Following the analysis mentioned above, the result data reveal fewer standard errors and estimates proposed variables to significant, and the proposed method is quite superior to the traditional algorithm in the recognition rate. The reason is that the CNN method has a strong dependence on data and contains prior knowledge of the pre-training model, and proposed framework can enhance the robustness of data preprocessing. The experimental results show that the selected hyperparameters that are suitable for iris recognition. These optimizations make the proposed framework have a better recognition effect.

\quad
\section{\rmfamily Conclusions}

A comprehensive overview of mobile environment devices and an implement of iris authentication system is presented, and the merits and drawbacks of the methods used in each study are analyzed. Some of researches operate on images captured using non-mobile neural network algorithms, resulting in high costs, low operation speed and complex calculations for mobile terminal devices. This investigations showed that the method of deep learning is developing towards a complex hierarchical structure. We develop a learning framework with multi-modalities for detection, extraction, normalization and recognition for high-resolution iris images in our iris authentication system. Considering the diversity of iris samples, the proposed Mask R-CNN architecture applies the local iris ground truth to achieve the robust iris localization efficiently, and the zero-padding layer can deal with the complex and varying scale-variant features flexibly. Thus the whole framework also shows the dynamic and flexible characteristics in the prepossessing. In addition, an approach of fined tuned-based Inception V4 architecture to recognize the iris is introduced, which is tested by 5-fold cross validation with a real data set. Our experiment results for the proposed framework indicates that the performance is better than the state-of-the-art methodologies in detection capability, recognition rate, and the robustness of the system.  
 
In consequence, qualitative and quantitative research designs were adopted to provide experimental results and attempts to explain the scale-variant features of models learning through iris recognition. The proposed solutions are suitable for high performance built-in GPU mobile devices and help researchers estimate analysis results for further research in the mobile environment.

\quad
\section*{\rmfamily Acknowledgments}
We thank the support of the Multimedia Department of Computer Science and Information Technology at the Putra Malaysia University, and the laboratory of Computer-Assisted Surgery and Diagnostic (CASD). This research was undertaken at CASD, and the support of CASD is gratefully acknowledged. I would like to thank all the contributors to this study for their prompt responses to different opinions, their continued support and encouragement and the vibrant research atmosphere that they have provided.

\quad
\section*{\rmfamily Author Contributions} 
\noindent \textbf{Siming Zheng:} Conceptualization, Data curation, Investigation, Methodology, Formal analysis, Visualization, Writing – original draft, Writing – review \& editing.
\quad\\
\noindent \textbf{Rahmita Wirza O.K. Rahmat:} Conceptualization, Project administration, Supervision, Funding acquisition, Writing \& original draft.
\quad\\
\noindent \textbf{Fatimah Khalid:} Formal analysis, Methodology, Validation, Formal analysis, Writing – review \& editing.
\quad\\
\noindent \textbf{Nurul Amelina Nasharuddin:} Data curation, Software, Resources, Visualization, Writing – review \& editing.
\quad\\

\quad
{

\end{document}
\begin{thebibliography}{67}

\sffamily{

\bibitem{ref1}Folorunso C. O., Asaolu O. S. \& Popoola O. P. A Review of Voice-Base Person Identification: State-of-the-Art. Covenant Journal of Engineering Technology (CJET) Vol.3 No.1, June 2019 ISSN: p 2682-5317 e 2682-5325. \href{https://doi.org/10.20370/2cdk-7y54}{https://doi.org/10.20370/2cdk-7y54}

 

\bibitem{ref2}Hui, D. O. Y., Yuen, K. K., Zahor, B. A. F. B. S. M., Wei, K. L. C., \& Zaaba, Z. F. (2018). An assessment of user authentication methods in mobile phones. \href{https://https://doi.org/10.1063/1.5055518}{https://doi.org/10.1063/1.5055518}
 
\bibitem{ref3}Chen, J., \& Ran, X. (2019). Deep Learning With Edge Computing: A Review. Proceedings of the IEEE, 1–20. \href{https://doi.org/10.1109/jproc.2019.2921977}{https://doi.org/10.1109/jproc.2019.2921977}
 

\bibitem{ref4}Noruzi, A., Mahlouji, M. \& Shahidinejad, A. Artif Intell Rev (2019). \href{https://doi.org/10.1007/s10462-019-09776-7}{https://doi.org/10.1007/s10462-019-09776-7}

\bibitem{ref5}Deng, Y. (2019). Deep learning on mobile devices: a review. In Mobile Multimedia/Image Processing, Security, and Applications 2019, vol. 10993, 109930A, International Society for Optics and Photonics.

\bibitem{ref6}ARROWSNX F-04G (2015). \href{http://www.fujitsu.com/global/about/resources/news/press-releases/2015/0525-01.html}{http://www.fujitsu.com/global/about/resour\\ces/news/press-releases/2015/0525-01.html} Accessed 15 October 2016.

\bibitem{ref7}HUAWEI P30 Pro (2019). Accessed October 2019. \href{https://consumer.huawei.com/en/phones/p30/specs/}{https://consumer.huawei.com/en/phones/p30/specs/} 

\bibitem{ref8}NVIDIA. Tegra K1 Next-Get Mobile Processor., 2014. \href{http://www.nvidia.com/object/tegra-k1- processor.html}{http://www.nvidia.com/object/tegra-k1- processor.html} Accessed Ovtober 2019.

\bibitem{ref9}Hess, Christopher David, "Design of an embedded iris recognition system for use with a multi-factor authentication system." (2019). Electrical Engineering Undergraduate Honors Theses. 60. \href{https://scholarworks.uark.edu/eleguht/60}{https://scholarworks.uark.edu/eleguht/60}

\bibitem{ref10}Gorodnichy, D. O., \& Chumakov, M. P. (2019). Analysis of the effect of ageing, age, and other factors on iris recognition performance using NEXUS scores dataset. IET Biometrics, 8(1), 29–39. \href{https://doi.org/10.1049/iet-bmt.2018.5105}{https://doi.org/10.1049/iet-bmt.2018.5105}


\bibitem{ref11}M. M. Khaladkar and S. R. Ganorkar, “A Novel Approach for Iris Recognition,” vol. 1, no. 4, 2012.

\bibitem{ref12}J. Daugman and C. Downing, Epigenetic randomness, complexity and singularity of human iris patterns, Proceedings of the Royal Society of London B: Biological Sciences, no. December 2000, pp. 1737–1740, 2001.

\bibitem{ref13}N. Ahmadi, M. Nilashi, Iris texture recognition based on multilevel 2-D haar wavelet decomposition and hamming distance approach, J. Soft Comput. Dec. Support Sys. 5 (3) (2018) 16–20.

\bibitem{ref14}N.Y. Tay, K. M. Mok, A review of iris recognition algorithms in information technology international symposium on vol. 2, pp. 1-7 Aug 2008.

\bibitem{ref15}Samant, P., Agarwal, R., \& Bansal, A. (2017). Enhanced discrete cosine transformation feature based iris recognition using various scanning techniques. 2017 4th IEEE Uttar Pradesh Section International Conference on Electrical, Computer and Electronics (UPCON). \href{https://doi.org/10.1109/upcon.2017.8251128}{https://doi.org/10.1109/upcon.2017.8251128}


\bibitem{ref16}Deshpande, B., \& Jayaswal, D. (2018). Fast and Reliable Biometric Verification System Using Iris. 2018 Second International Conference on Inventive Communication and Computational Technologies (ICICCT). \href{https://doi.org/10.1109/icicct.2018.8473300}{https://doi.org/10.1109/icicct.2018.8473300}

\bibitem{ref17}J. G. Daugman, “How iris recognition works,” IEEE Trans. Circuits Syst. Video Technol., vol. 14, no. 1, pp. 21–30, Jan. 2004.

\bibitem{ref18}Mohammed Hamzah Abed. Iris recognition model based on Principal Component analysis and 2 level Haar wavelet transform: Case study CUHK and UTIRIS iris databases. Journal of College of Education/Wasit, VL-27, 2017.
 
\bibitem{ref19}Kaudki Onkar \& Bhurchandi Kishor. (2018). A Robust Iris Recognition Approach Using Fuzzy Edge Processing Technique. 2018 9th International Conference on Computing, Communication and Networking Technologies (ICCCNT). \href{https://doi.org/10.1109/icccnt.2018.8493855}{https://doi.org/10.1109/icccnt.2018.8493855}
 

\bibitem{ref20}K. W. Bowyer, S. E. Baker, A. Hentz, K. Hollingsworth, T. Peters, and P. J. Flynn, “Factors that degrade the match distribution in iris biometrics,” Identity in the Information Society, vol. 2, no. 3, pp. 327–343, 2009.

\bibitem{ref21}Bowyer, K. W., Baker, S. E., Hentz, A., Hollingsworth, K., Peters, T., \& Flynn, P. J. (2009). Factors that degrade the match distribution in iris biometrics. Identity in the Information Society, 2(3), 327–343. \href{https://doi.org/10.1007/s12394-009-0037-z}{https://doi.org/10.1007/s12394-009-0037-z}

\bibitem{ref22}E. Garea, J.M. Colores, M.S. García , L.M. Zamudio , A .A . Ramírez, Cross-sensor Iris verification applying robust fused segmentation algorithms, in: IEEE. Pro- ceedings of International Conference on Biometrics. ICB 2015, 2015, pp. 17–22.

\bibitem{ref23}Llano, E. G., García Vázquez, M. S., Vargas, J. M. C., Fuentes, L. M. Z., \& Ramírez Acosta, A. A. (2018). Optimized robust multi-sensor scheme for simultaneous video and image iris recognition. Pattern Recognition Letters, 101, 44–51. \href{https://doi.org/10.1016/j.patrec.2017.11.012}{https://doi.org/10.1016/j.patrec.2017.11.012}


\bibitem{ref24}Sunil Chawla and Aashish Oberoi, “A Robust Algorithm for Iris Segmentation and Normalization using Hough Transform,” Global Journal of Business Management and Information Technology, Volume 1, Number 2 (2011), pp. 69-76, 2011.

\bibitem{ref25}Anand Deshpande, Prashant Patavardhan, “Segmentation And Quality Analysis Of Long Range Captured Iris Image,” ICTACT Journal on Image and Video Processing, 2016.

\bibitem{ref26}Reddy, N., Rattani, A., \& Derakhshani, R. (2016). A robust scheme for iris segmentation in mobile environment. 2016 IEEE Symposium on Technologies for Homeland Security (HST). \href{https://doi.org/10.1109/ths.2016.7568948}{https://doi.org/10.1109/ths.2016.7568948}

\bibitem{ref27}Bansal, R., Juneja, A., Agnihotri, A., Taneja, D., \& Sharma, N. (2017). A fuzzfied approach to Iris recognition for mobile security. 2017 International Conference of Electronics, Communication and Aerospace Technology (ICECA).   \href{https://doi.org/10.1109/iceca.2017.8203680}{https://doi.org/10.1109/iceca.2017.8203680}


\bibitem{ref28}McCabe, T. J., \& Butler, C. W. (1989). Design complexity measurement and testing. Communications of the ACM, 32(12), 1415–1425. \href{https://doi.org/10.1145/76380.76382}{https://doi.org/10.1145/76380.76382}
 
\bibitem{ref29}Mrinal Kanti Debbarma, Swapan Debbarma, and Nikhil Debbarma, "A Review and Analysis of Software Complexity Metrics in Structural Testing," International Journal of Computer and Communication Engineering vol. 2, no. 2, pp. 129-133, 2013.

\bibitem{ref30}Dong, H., Supratak, A., Mai, L., Liu, F., Oehmichen, A., Yu, S., \& Guo, Y. (2017). TensorLayer: A Versatile Library for Efficient Deep Learning Development. Proceedings of the 2017 ACM on Multimedia Conference - MM ’17. \href{https://doi.org/10.1145/3123266.3129391}{https://doi.org/10.1145/3123266.3129391}


\bibitem{ref31}Galdi, C., \& Dugelay, J.-L. (2016). Fusing iris colour and texture information for fast iris recognition on mobile devices. 2016 23rd International Conference on Pattern Recognition (ICPR). \href{https://doi.org/10.1109/icpr.2016.7899626}{https://doi.org/10.1109/icpr.2016.7899626}

\bibitem{ref32}Falconer, Kenneth (1990). Fractal geometry: mathematical foundations and applications. Chichester: John Wiley. pp. 38–47.

\bibitem{ref33}Sarkar, N., \& Chaudhuri, B. B. (1994). An efficient differential box-counting approach to compute fractal dimension of image. IEEE Transactions on Systems, Man, and Cybernetics, 24(1), 115–120. \href{https://doi.org/10.1109/21.259692}{https://doi.org/10.1109/21.259692}

\bibitem{ref34}R. Dobrescu, D. Popescu, “Image processing applications based on texture and fractal analysis,” Applied Signal and Image Processing: Multidisciplinary Advancements, pp. 226–250, 2011.

\bibitem{ref35}D. Popescu, L. Ichim, T. Caramihale, “Texture based method for automated detection, localization and evaluation of the exudates in retinal images,” 22nd International Conference on Neural Information Processing (ICONIP), S. Arik et al. (Eds.): Neural Information Processing, ICONIP 2015, Part IV, LNCS 9492, pp. 463-472, 2015. 

\bibitem{ref36}Nayak, S. R., Mishra, J., \& Jena, P. mohan. (2017). Fractal analysis of image sets using differential box counting techniques. International Journal of Information Technology, 10(1), 39–47. \href{https://doi.org/10.1007/s41870-017-0062-3}{https://doi.org/10.1007/s41870-017-0062-3}

\bibitem{ref37}Lamiaa A. Elrefaei, Doaa H. Hamid, Afnan A. Bayazed1 \& Sara S. Bushnak1 and Shaikhah Y. Maasher. “Developing Iris Recognition System for Smartphone Security,” Springer, July 2017.

\bibitem{ref38}Hajari, K., \& Bhoyar, K. (2015). A review of issues and challenges in designing Iris recognition Systems for noisy imaging environment. 2015 International Conference on Pervasive Computing (ICPC). \href{https://doi.org/10.1109/pervasive.2015.7087003}{https://doi.org/10.1109/pervasive.2015.7087003}


\bibitem{ref39}Yooyoung Lee, Ross J. Micheals, James J. Filliben and P. Jonathon Phillips, “Vasir: An Open-Source Research Platform for Advanced Iris Recognition Technologies”, Journal of Research of the National Institute of Standards and Technology, Vol. 118, pp. 218-259, 2013.

\bibitem{ref40}Zhaofeng He, Tieniu Tan, Zhenan Sun and Xianchao Qiu, “Robust Eyelid, Eyelash and Shadow Localization for Iris Recognition”, Proceedings of 15th IEEE International Conference on Image Processing, pp. 265-268, 2008.

\bibitem{ref41}Kevin W. Bowyer, Karen Hollingsworth and Patrick J. Flynn, “Image Understanding for Iris Biometrics: A Survey”, Computer Vision and Image Understanding, Vol. 110, No. 2, pp. 281-307, 2008.

\bibitem{ref42}Kaiming He, Georgia Gkioxari, Piotr Dollar, Ross Girshick. Mask R-CNN.The IEEE. International Conference on Computer Vision (ICCV), 2017, pp. 2961-2969.

\bibitem{ref43}Ross Girshick; The IEEE International Conference on Computer Vision (ICCV), 2015, Fast R-CNN. pp. 1440-1448.

\bibitem{ref44}[J] Mahdi S. Hosseini, Babak N. Araabi and H. Soltanian-Zadeh, Pigment Melanin: Pattern for Iris Recognition, IEEE Transactions on Instrumentation and Measurement, vol.59, no.4, pp.792-804, April 2010. \href{https://doi.org/10.6084/m9.figshare.10279592}{https://doi.org/10.6084/m9.figshare.10279592}

\bibitem{ref45}Tsung-Yi Lin, Michael Maire, Serge Belongie, James Hays, Pietro Perona, Deva Ramanan, Piotr Dollár, and C Lawrence Zitnick. Microsoft COCO: Common objects in context. In European
Conference on Computer Vision, pp. 740–755, 2014.

\bibitem{ref46}Aginako, N., Castrillón-Santana, M., Lorenzo-Navarro, J., Martínez-Otzeta, J. M., \& Sierra, B. (2017). Periocular and iris local descriptors for identity verification in mobile applications. Pattern Recognition Letters, 91, 52–59. \href{https://doi.org/10.1016/j.patrec.2017.01.021}{https://doi.org/10.1016/j.patrec.2017.01.021}
 

\bibitem{ref47}Padole, C. N., \& Proenca, H. (2012). Periocular recognition: Analysis of performance degradation factors. 2012 5th IAPR International Conference on Biometrics (ICB). \href{https://doi.org/10.1109/icb.2012.6199790}{https://doi.org/10.1109/icb.2012.6199790}


\bibitem{ref48}Anis Farihan Mat Raffei, Tole Sutikno, Hishammuddin Asmuni, Rohayanti Hassan, Razib M Othman, Shahreen Kasim, Munawar A Riyadi. Fusion Iris and Periocular Recognitions in Non-Cooperative Environment. IJEEI, 2019. \href{https://doi.org/10.11591/ijeei.v7i3.1147}{https://doi.org/10.11591/ijeei.v7i3.1147}

\bibitem{ref49}Vincent Dumoulin and Francesco Visin. A guide to convolution arithmetic for deep learning. arXiv preprint \href{https://arxiv.org/abs/1603.07285}{arXiv:1603.07285}, 2016.

\bibitem{ref50}Montanari L, Basu B, Spagnoli A, Broderick BM. A padding method to reduce edge effects for enhanced damage identification using wavelet analysis. Mech Syst Signal Pr. 2015;52-53:264-277.

\bibitem{ref51}Tang X., Xie J., Li P. (2017) Deep Convolutional Features for Iris Recognition. In: Zhou J. et al. (eds) Biometric Recognition. CCBR 2017. Lecture Notes in Computer Science, vol 10568. Springer, Cham.

\bibitem{ref52}Ribeiro, E., Uhl, A., \& Alonso-Fernandez, F. (2018). Iris Super-Resolution using CNNs: is Photo-Realism Important to Iris Recognition? IET Biometrics. \href{https://doi.org/10.1049/iet-bmt.2018.5146}{https://doi.org/10.1049/iet-bmt.2018.5146}


\bibitem{ref53}Szegedy C, Ioffe S, Vanhoucke V, et al. Inception-v4, Inception-ResNet and the Impact of Residual Connections on Learning[J]. 2016.

\bibitem{ref54}C. Szegedy, V. Vanhoucke, S. Ioffe, J. Shlens, and Z. Wojna. Rethinking the inception architecture for computer vision. In CVPR, 2016:2818-2826.

\bibitem{ref55}Siyu Huang, Xi Li, Zhiqi Cheng, Zhongfei Zhang, and Alexander G. Hauptmann. Stacked pooling: Improving crowd counting by boosting scale invariance. CoRR, abs/1808.07456, 2018.


\bibitem{ref56}Cui, Y., Zhou, F., Wang, J., Liu, X., Lin, Y., \& Belongie, S. (2017). Kernel Pooling for Convolutional Neural Networks. 2017 IEEE Conference on Computer Vision and Pattern Recognition (CVPR). \href{https://doi.org/10.1109/cvpr.2017.325}{https://doi.org/10.1109/cvpr.2017.325}
 

\bibitem{ref57}Smith, L. N. (2017). Cyclical Learning Rates for Training Neural Networks. 2017 IEEE Winter Conference on Applications of Computer Vision (WACV). \href{https://doi.org/10.1109/wacv.2017.58}{https://doi.org/10.1109/wacv.2017.58}
 

\bibitem{ref58} Ruder, S. An overview of gradient descent optimization algorithms. CoRR, abs/1609.04747, 2016.

\bibitem{ref59}Wilson, A. C, Roelofs, R., Stern, M., Srebro, N., and Recht, B. The marginal value of adaptive gradient methods in machine learning. arXiv preprint \href{https://arxiv.org/abs/1705.08292}{arXiv:1705.08292}, 2017.


\bibitem{ref60}I. Loshchilov and F. Hutter. Fixing weight decay regularization in adam. arXiv preprint \href{https://arxiv.org/abs/1711.05101}{arXiv:1711.05101}, 2017.

\bibitem{ref61} Reddi, Sashank J, Kale, Satyen, and Kumar, Sanjiv. On the convergence of adam and beyond. In International Conference on Learning Representations (ICLR), 2018.

\bibitem{ref62}Kohavi, R., et al. 1995. A study of cross-validation and bootstrap for accuracy estimation and model selection. In IJCAI, volume 14.

\bibitem{ref63}Yadav, S., \& Shukla, S. (2016). Analysis of k-Fold Cross-Validation over Hold-Out Validation on Colossal Datasets for Quality Classification. 2016 IEEE 6th International Conference on Advanced Computing (IACC). \href{https://doi.org/10.1109/iacc.2016.25}{https://doi.org/10.1109/iacc.2016.25}
 

\bibitem{ref64}N. Corp., “Jetson nano,” \href{https://developer.nvidia.com/embedded/buy/jetsonnano-devkit}{developer.nvidia.com/embedded/buy/jetsonnano-devkit}, 2019, [Online; accessed 09/27/19].

\bibitem{ref65}Ramyad Hadidi, Jiashen Cao, Yilun Xie, Bahar Asgari, Tushar Krishna, Hyesoon Kim. Characterizing the Deployment of Deep Neural Networks on Commercial Edge Devices. IEEE International Symposium on Workload Characterization (IISWC), Orlando, Florida (2019).

\bibitem{ref66}Wong, T.T.; Yeh, P.Y. Reliable Accuracy Estimates from k-fold Cross Validation. IEEE Trans. Knowl. Data Eng. 2019.

\bibitem{ref67}Leslie N Smith. A disciplined approach to neural network hyper-parameters: Part 1–learning
rate, batch size, momentum, and weight decay. arXiv preprint \href{https://arxiv.org/abs/1803.09820}{arXiv:1803.09820}, 2018.

}


\end{thebibliography}
